\documentclass[10pt,twocolumn,letterpaper]{article}

\usepackage{iccv}
\usepackage{times}
\usepackage{epsfig}
\usepackage{graphicx}
\usepackage{amsmath}
\usepackage{amssymb}
\usepackage{booktabs}
\usepackage{algorithm}
\usepackage{algpseudocode}
\usepackage[table,xcdraw,dvipsnames]{xcolor}
\usepackage{soul}
\usepackage[export]{adjustbox}
\usepackage[accsupp]{axessibility}  

\usepackage[pagebackref=true,breaklinks=true,letterpaper=true,colorlinks,bookmarks=false]{hyperref}

\usepackage[capitalize]{cleveref}
\crefname{section}{Section}{Sections}
\Crefname{section}{Section}{Sections}
\Crefname{table}{Table}{Tables}
\crefname{table}{Tabla}{Tables}

\iccvfinalcopy 

\def\iccvPaperID{8958}

\ificcvfinal\fi

\begin{document}

\title{H3WB: Human3.6M 3D WholeBody Dataset and Benchmark}

\author{Yue Zhu
\and 
Nermin Samet
\and
David Picard
\and
LIGM, Ecole des Ponts, Univ Gustave Eiffel, CNRS, Marne-la-Vallée, France\\
{\tt\small \{yue.zhu, nermin.samet, david.picard\}@enpc.fr} \\
\small{Code and dataset:  \url{https://github.com/wholebody3d/wholebody3d}}
}

\maketitle
\ificcvfinal\thispagestyle{empty}\fi

\begin{abstract}

We present a benchmark for 3D human whole-body pose estimation, which involves identifying accurate 3D keypoints on the entire human body, including face, hands, body, and feet. Currently, the lack of a fully annotated and accurate 3D whole-body dataset results in deep networks being trained separately on specific body parts, which are combined during inference. Or they rely on pseudo-groundtruth provided by parametric body models which are not as accurate as detection based methods. To overcome these issues, we introduce the Human3.6M 3D WholeBody (H3WB) dataset, which provides whole-body annotations for the Human3.6M dataset using the COCO Wholebody layout. H3WB comprises 133 whole-body keypoint annotations on 100K images, made possible by our new multi-view pipeline. We also propose three tasks: i) 3D whole-body pose lifting from 2D complete whole-body pose, ii) 3D whole-body pose lifting from 2D incomplete whole-body pose, and iii) 3D whole-body pose estimation from a single RGB image. Additionally, we report several baselines from popular methods for these tasks. Furthermore, we also provide automated 3D whole-body annotations of TotalCapture and experimentally show that when used with H3WB it helps to improve the performance.

\end{abstract}

\begin{figure*}[tb]
  \centering
  \includegraphics[width=0.9\linewidth]{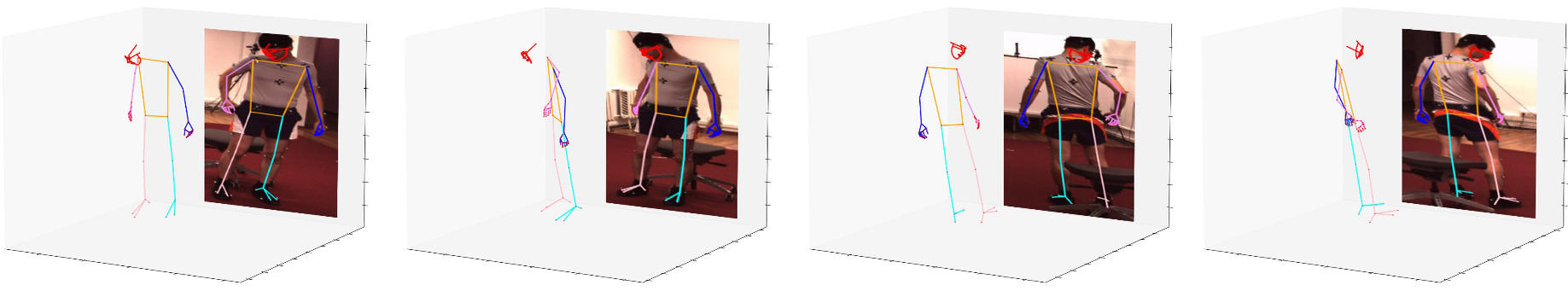}
  \caption{The H3WB dataset has 133 whole-body keypoint annotations in 3D as well as their respective projections in 2D.}
  \label{fig:general-idea}
\end{figure*}


\section{Introduction}
\label{sec:intro}

3D Human pose estimation is the task of localizing human body keypoints in images which is critical to analyze human behavior, expressions, emotions, intentions, and how people communicate and interact with the physical world. As a result, 3D human pose estimation has an important role in several vision tasks and applications such as robotics~\cite{gu2019home, garcia2019human,gui2018teaching} or augmented/virtual reality~\cite{mehta2018single, bagautdinov2021driving, wang2021scene, Zanfir19aaai}.
However, to make more accurate predictions about human behaviors, we need more than a few body keypoints. To that end, 3D whole-body pose estimation aims to detect face, hand and foot keypoints in addition to the standard human body keypoints of classical 3D human pose estimation.

The lack of accurate 3D datasets has made 3D whole-body pose estimation a challenging task, leading previous works to focus on separate body parts and train separate models on different datasets for 3D body pose~\cite{arnab2019exploiting, habibie2019wild, mehta2018single, moon2019camera, lcrnet, bagautdinov2021driving, kanazawa2018end, kolotouros2019learning, kolotouros2019convolutional, rong2022chasing, wang2020motion, xiang2019monocular, Zanfir20eccv, DBLP:journals/corr/abs-2001-02024}, 3D hand pose~\cite{cai2018weakly, mueller2018ganerated, zimmermann2017learning, boukhayma20193d, ge20193d, iqbal2018hand, zhang2019end, hampali2021handsformer}, or 3D face landmarks~\cite{sanyal2019learning, cao2018pose, chang2018expnet}. However, directly ensembling separate body part models during inference suffers from issues arising from datasets' biases, pose and scales, and complex inference pipelines. Distillation from pretrained models has been used to overcome these issues, with FrankMocap~\cite{franmocap} using three specialized pretrained models to estimate 65 whole-body keypoints (22 on the body, 40 on the hands, and 3 on the face), and DOPE~\cite{dope} also using three specialized models to output 139 whole-body keypoints (13 on the body, 42 on the hands, and 84 on the face).

Alternatively, parametric body models can be fitted to obtain whole-body pose, as has been proposed in ExPose~\cite{expose}, SMPLify-X~\cite{smplifyx} or Monocular Total Capture (MTC)~\cite{xiang2019monocular}. 
While parametric models enable sampling an almost infinite number of keypoints from the mesh~\cite{afit,fieraru2020three,fieraru2021learning}, their accuracy is usually less than that of detection based methods on fine body parts like hands and feet (see supplementary material for examples from the literature).
Indeed, parametric models are tailored for visual applications such as realistic motion generation~\cite{petrovich22temos} or avatar capture~\cite{saitoCVPR2021}, where a realistic capture of the body shape and pose is more important than fine grain accuracy. Relying on a small set of data-driven pose parameters ensures realism but also limits their flexibility in representing complex unusual poses. For applications where the body shape is not needed but the accuracy of the keypoints is essential, like high performance sport analysis or ergonomics, detection based methods are thus the preferred solution.

Furthermore, 3D whole-body pose estimation has not been fully explored in the literature due to the absence of a representative and accurate benchmark. As previously mentioned, existing 3D whole-body methods either rely on specific datasets and models for different body parts, leading to complex training pipelines and heterogeneous evaluations, or utilize parametric models that prioritize shape capture over highly precise keypoints. In addition, unified methods vary significantly in terms of keypoint layout definition, number of keypoints and distribution of keypoints across body parts (see Table~\ref{tab:statistic}). These significant dataset disparities and the absence of a standard benchmark make it challenging to compare methods fairly.

To address the above issues, we propose a new large-scale dataset for accurate 3D whole-body pose estimation called Human3.6M 3D WholeBody, or H3WB for short (see ~\autoref{fig:general-idea}). Our dataset extends  Human3.6M~\cite{journals/pami/IonescuPOS14(H36m),IonescuSminchisescu11} with 3D whole-body keypoint annotations. It consists of 133 paired 2D and 3D whole-body keypoint annotations for a set of 100k images from Human3.6M, following the same layout used in COCO WholeBody~\cite{jin2020whole(COCO-WholeBody)}. More specifically, in addition to the standard 17 body keypoints, the dataset has 42 hand keypoints, 6 foot keypoints and 68 facial landmarks. H3WB was automatically created in a 3 step process: We obtain an initial set of 3D annotations using multi-view geometry. Then, we trained a masked auto-encoder to complete the initial annotations. Finally, we refine the whole-body keypoints via a diffusion model. 
A manual annotation of 80K keypoints from 600 images shows our labels have an average error of 17mm which suggest the H3WB keypoints are very accurate for such a complex task.
We propose 3 tasks and benchmarks on the H3WB dataset for which we provide baselines: i) \textit{ 3D whole-body pose lifting from a complete 2D whole-body keypoints}, ii) \textit{3D whole-body pose lifting from incomplete 2D whole-body keypoints} (i.e. 2D whole-body with missing keypoints, which is more realistic), and iii) \textit{3D whole-body pose estimation from a single RGB image}.

\begin{table}[b]
    \scriptsize
    \centering
    \begin{tabular}{c c |c | c c c}
    \toprule
    Dataset & Size & Keypoints & Body & Hand & Face \\
    \midrule
    Human3.6M\cite{journals/pami/IonescuPOS14(H36m)} & 3.6M & 17 & 17 & &  \\
    3DPW\cite{Marcard_2018_ECCV(3DPW)} & 51k & 24 & 24 & &   \\
    LSP\cite{inproceedings(LSP)} & 10k & 14 &  14 & &  \\
    3DHP\cite{Marcard_2018_ECCV(3DPW)} & $>$1.3M & 17  & 17 & &  \\
    Panoptic\cite{Joo_2015_ICCV(Panoptic)} & 1.5M & 15  & 15 & &  \\
    MTC\cite{xiang2019monocular} & 834K & 20 & 20 & & \\
    \midrule
    InterHand2.6M\cite{DBLP:journals/corr/abs-2008-09309(hand3)} & 2.6M & 21  & & 21 &  \\
    FreiHAND\cite{DBLP:journals/corr/abs-1909-04349(hand5)} & 37k & 21 & & 21 &  \\
    RHD\cite{zimmermann2017learning} & 44K & 21 & & 21 &  \\
    MTC\cite{xiang2019monocular} & 111K & 21 & & 21 & \\
    \midrule
        TotalCapture\cite{DBLP:journals/corr/abs-1801-01615(totalcapture)} & 1.9M & 127 & 21 & 16+16 & 74 \\
        
         ExPose\cite{expose} & 33K & 144  & 25 & 15+15 & 89 \\
    \midrule
    H3WB & 100k & 133 & 23 & 21+21 & 68 \\
    \bottomrule
    \end{tabular}
    
    \caption{Overview of datasets for 3D human pose estimation.}
    \label{tab:statistic}
\end{table}

Our contributions can be summarized as follows. 1) We propose a method to create detailed 3D human pose keypoints from multi-view images. 2) We propose H3WB, the first accurate public benchmark dataset for 3D whole-body pose estimation, using the aforementioned method. 
Our benchmark can easily leverage existing results in 2D and enables the community to build upon existing high-quality 2D detectors on COCO. Unifying the 3D whole-body pose estimation with the COCO 2D benchmark will greatly benefit the research community. 
3) We provide baselines for the 3 tasks of H3WB, which we believe will encourage the community to explore 3D whole-body pose estimation more and accelerate progress in the field. 4) Additionally, we provide 3D whole-body annotations for the TotalCapture~\cite{DBLP:journals/corr/abs-1801-01615(totalcapture)} dataset, and show that when combined with the H3WB dataset it improves the performance of pose lifting tasks.

\section{Related work}

\noindent\textbf{3D Body, hand and face pose estimation.} There are two main groups of prominent approaches in 3D human pose estimation. The first group directly estimates 3D body pose from a single RGB image~\cite{DBLP:journals/corr/PavlakosZDD16(CoarsetoFine),mehta2017monocular(3DHP),mehta2018single,moon2019camera,DBLP:journals/corr/PavlakosZDD16(CoarsetoFine),lcrnet}. The second group follows two stage approach where they first localize 2D  keypoints and then lift 2D human pose to 3D space~\cite{DBLP:journals/corr/MartinezHRL17, DBLP:journals/corr/abs-2011-14679(canonpose), Iqbal_2020_CVPR(MultiviewInTheWild),DBLP:journals/corr/BogoKLG0B16(SMPLifyModel),lassner2017unite,DBLP:journals/corr/MartinezHRL17}. Several optimization based methods~\cite{DBLP:journals/corr/BogoKLG0B16(SMPLifyModel),lassner2017unite}, utilize 2D keypoints to initialize a parametric model of the human body such as SMPL~\cite{SMPL:2015(SMPLModel)}. Several works attempt to eliminate the requirement of 3D annotations using 2D multi-view supervision to estimate 3D human pose~\cite{DBLP:journals/corr/abs-2011-14679(canonpose), Xu_2020_CVPR(Kinematic)} or temporal supervision with video~\cite{Choi_2021_CVPR(TempConsistency, Liu_2020_CVPR(PoseReconstruction)}. 
3D hand pose estimation methods share similar approaches as the body counterparts. First group of works, estimates hand pose from a single RGB image by directly regressing 3D hand keypoints~\cite{yang2019aligning}, mesh vertices ~\cite{ge20193d,kulon2020weakly}, and parameters of parametric 3D hand models such as MANO~\cite{boukhayma20193d,romero2022embodied,baek2019pushing,chen2021camera,chen2021model,zhang2019end}. Second groups of works rely on intermediate 2D representations such as 2D keypoints and feature maps~\cite{cai2018weakly,mueller2018ganerated,zhang2019end,zimmermann2017learning,DBLP:journals/corr/abs-2005-04551(hand2)}. Similarly, predominant 3D face pose estimation methods regress the dense 3D face landmarks~\cite{crispell2017pix2face,feng2018joint,jackson2017large}  and face model parameters~\cite{chang2018expnet,sanyal2019learning,tuan2017regressing,DBLP:journals/corr/abs-1903-08527(face2),DBLP:journals/corr/abs-2110-04800(face4)} based on 3DMM~\cite{blanz1999morphable}.

\noindent\textbf{3D Whole-body pose estimation.} There are a several methods~\cite{smplifyx, xiang2019monocular, zhou2021monocular,expose,dope,franmocap} jointly estimating 3D whole-body pose. The first group of works is based on parametric human body models such as Adam~\cite{DBLP:journals/corr/abs-1801-01615(totalcapture)} and SMPL-X~\cite{smplifyx}. MTC~\cite{xiang2019monocular} is based on the Adam model~\cite{DBLP:journals/corr/abs-1801-01615(totalcapture)}, and first gets 2.5D predictions, then optimizes the parameters of Adam. SMPLify-X optimizes the parameter of the SMPL-X model~\cite{smplifyx} to fit it to 2D keypoints. As a major drawback, optimization-based methods are relatively slow and highly sensitive to parameter initializations. Non-parametric methods~\cite{dope,expose, franmocap} follow different approaches to avoid heavy optimization procedure. DOPE~\cite{dope} and FrankMocap~\cite{franmocap} first train separate body, hand, and face models. Next, they combine those models within a learning framework. DOPE~\cite{dope} curates pseudo-ground truths from separate body models and uses those ground-truths to supervise the distillation model. Similar to DOPE, ExPose~\cite{expose} first obtains a pseudo-ground truth dataset by fitting SMPL-X model on in-the-wild images, and trains a joint model to output whole-body poses.  All these methods utilize many part-based datasets. Moreover, all output different whole-body layouts with a different numbers of whole-body keypoint. FrankMocap, DOPE, and SMPLify-X estimate whole-body pose with 65, 139 and 144 keypoints, respectively.

\noindent\textbf{Pose completion} completes a partially estimated pose by localizing missing keypoints. Carissimi \etal~\cite{DBLP:conf/eccv/CarissimiRBM18(completion1)} propose a denoising variational autoencoder network to fill the missing keypoints in 2D pose completion. Bautembach \etal~\cite{article(completion2)} selects a small subset of poses from a database based on their distance to an incomplete 3D pose, and replaces the missing keypoints with the corresponding averaged keypoints in the subset. Despite being critical for real-world scenarios, pose completion has not been sufficiently explored due to the lack of annotated keypoint datasets. Our 3D whole-body dataset can facilitate more exploration of 3D pose estimation from 2D incomplete human poses.

\section{The H3WB dataset}

In this section, we describe the making of the H3WB dataset\footnote{We consider the feet keypoints as a part of the body.}. Our objective is to build a keypoint based 3D whole-body dataset  including keypoints on the body, the face and the hands, and propose a benchmark. We use the same keypoint layout as COCO WholeBody dataset~\cite{jin2020whole(COCO-WholeBody)} with 133 keypoints. 

To that end, we build on the widely used Human3.6M dataset~\cite{journals/pami/IonescuPOS14(H36m)} for which we provide 3D whole-body keypoints. The H3WB building process is as follows: First, we use an off-the-shelf 2D whole-body detector combined with multi-view reconstruction to obtain an initial set of incomplete 3D whole-body keypoints. Next, we implement a completion network to fill in the keypoints missed by the multi-view geometric approach. Then, we develop a refinement method for the hands and the face to obtain more accurate keypoints. Finally, we perform quality assessment to select 25k 3D whole-body poses with high confidence and the 100k associated images from 4-view.

\subsection{Initial 3D whole-body dataset with OpenPifPaf}

We run the 2D whole-body detector from OpenPifPaf~\cite{kreiss2021openpifpaf} on all the 4 views from the training set of Human3.6M (S1, S5, S6, S7 and S8, 1 image per 5 frames). 
Since the cameras of Human3.6M are well calibrated, we can reconstruct keypoints in 3D  using standard multi-view geometry.

\begin{figure}
  \centering
  \includegraphics[width=\linewidth]{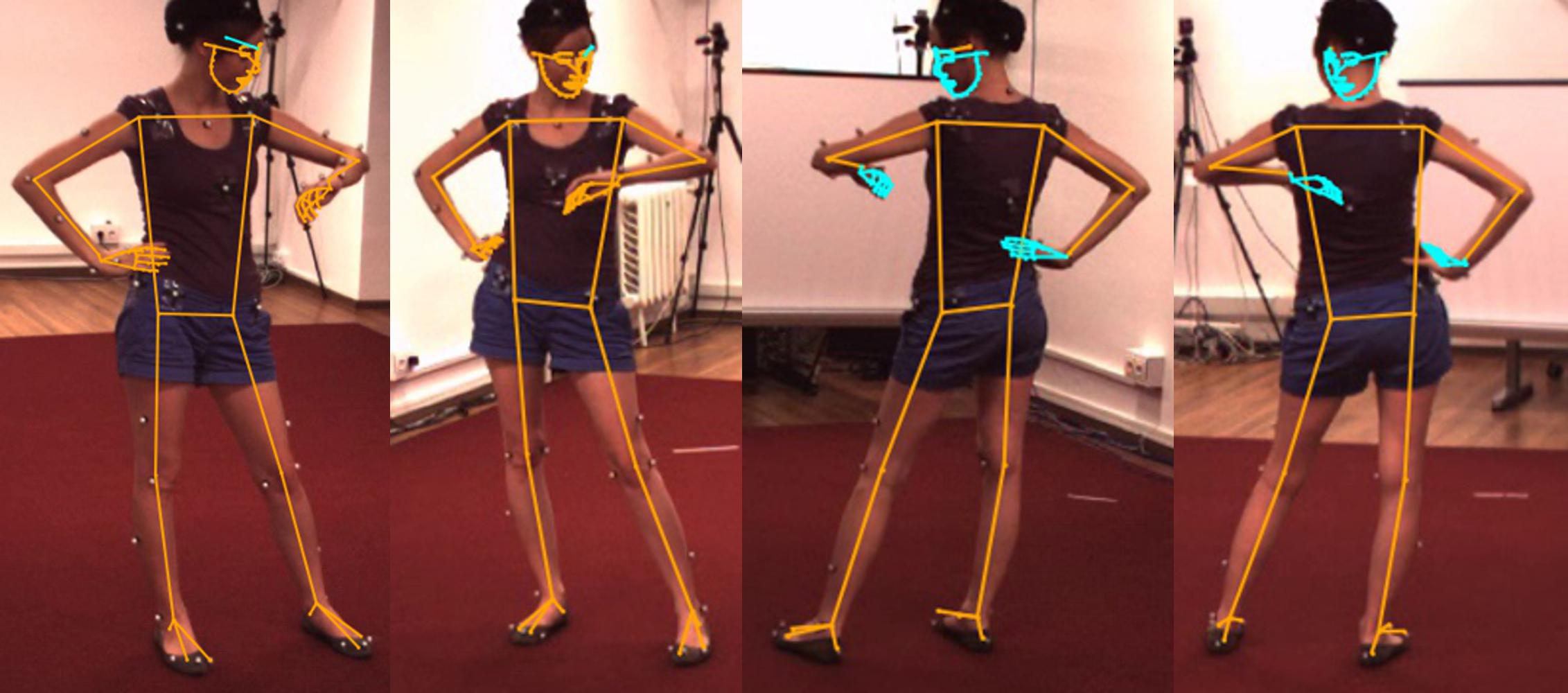}
  \caption{OpenPifPaf detects most of the non-occluded keypoints inside the image (orange keypoints). The occluded or undetected keypoints (cyan keypoints) are reprojections after 3D multi-view reconstruction. Notice that these reprojections do not always align with the images, like the right hand in the last view, which is probably due to OpenPifPaf not being perfectly accurate.}
  \label{fig:after-openpifpaf}
\end{figure}

The OpenPifPaf 2D whole-body detector can miss keypoints due to self-occlusions (hands, feet) or unfavorable camera viewpoints (facial landmarks). However, the four-view setup allows us to recover missing keypoints and obtain a complete 3D whole-body pose, provided each keypoint appears in at least two non-opposing views. An example of this process is shown in \autoref{fig:after-openpifpaf}. Using this method, we obtained 11,426 fully complete 3D whole-body poses with all 133 keypoints and 26,333 incomplete 3D whole-body poses where all keypoints appear in at least one view, resulting in a total of 37,759 3D whole-body poses with each keypoint appearing in at least one view.

We did not rely of the video information because the reconstruction problem becomes significantly more difficult when there is a motion between two frames. In the absence of motion, an additional frame does not help solve the occlusion problem. The results of our study demonstrate that multi-view labeling is sufficiently effective for our task (see \autoref{tab:step_score}, ``Geometry'' line).

\begin{figure}[tb]
  \centering
  \includegraphics[width=\linewidth]{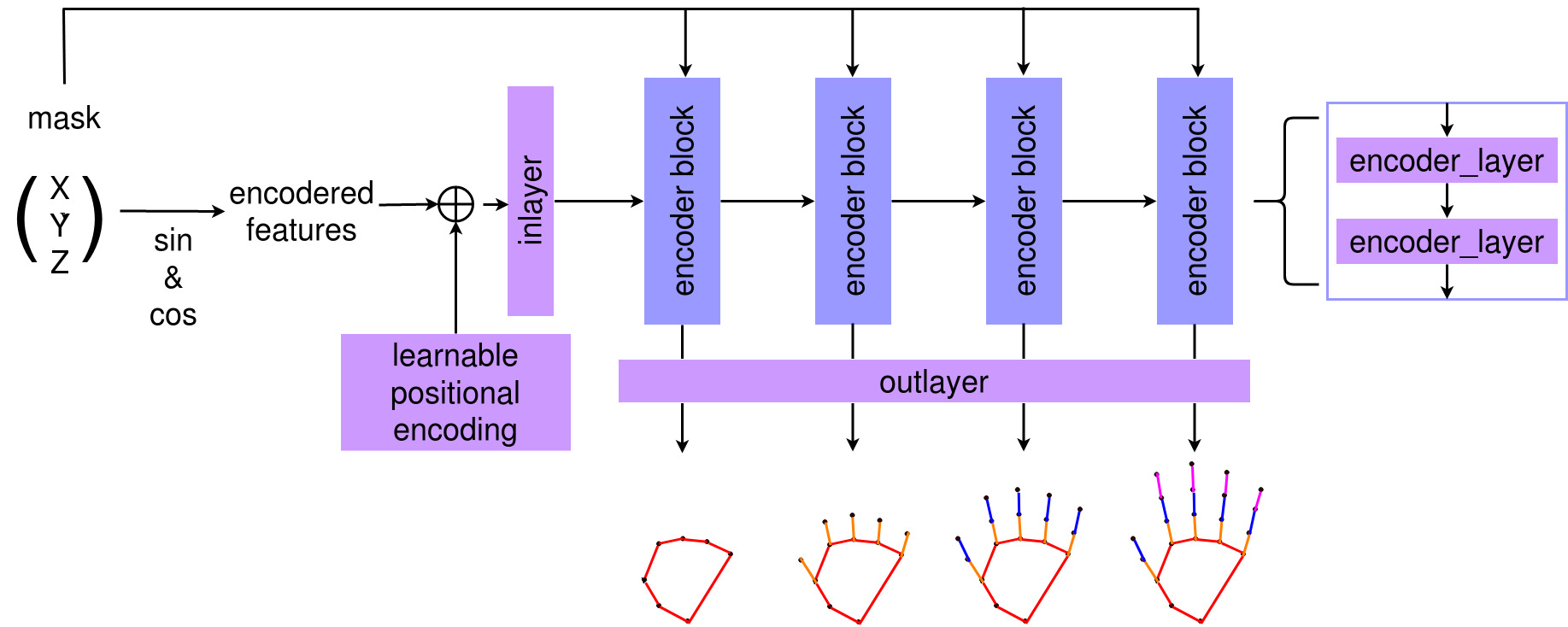}
  \caption{The completion network consists of one linear input layer, 4 transformer encoder blocks (each of them containing 2 transformer encoder layer with $d\_model=64$ and $n\_head=1$), and a linear output layer. At the end of each encoder block, the features are decoded by the output layer into a predicted position in a curriculum way where later blocks decode more keypoints.}
  \label{fig:completionnet}
\end{figure}

\begin{figure}[tb]
  \centering
  \includegraphics[width=0.85\linewidth]{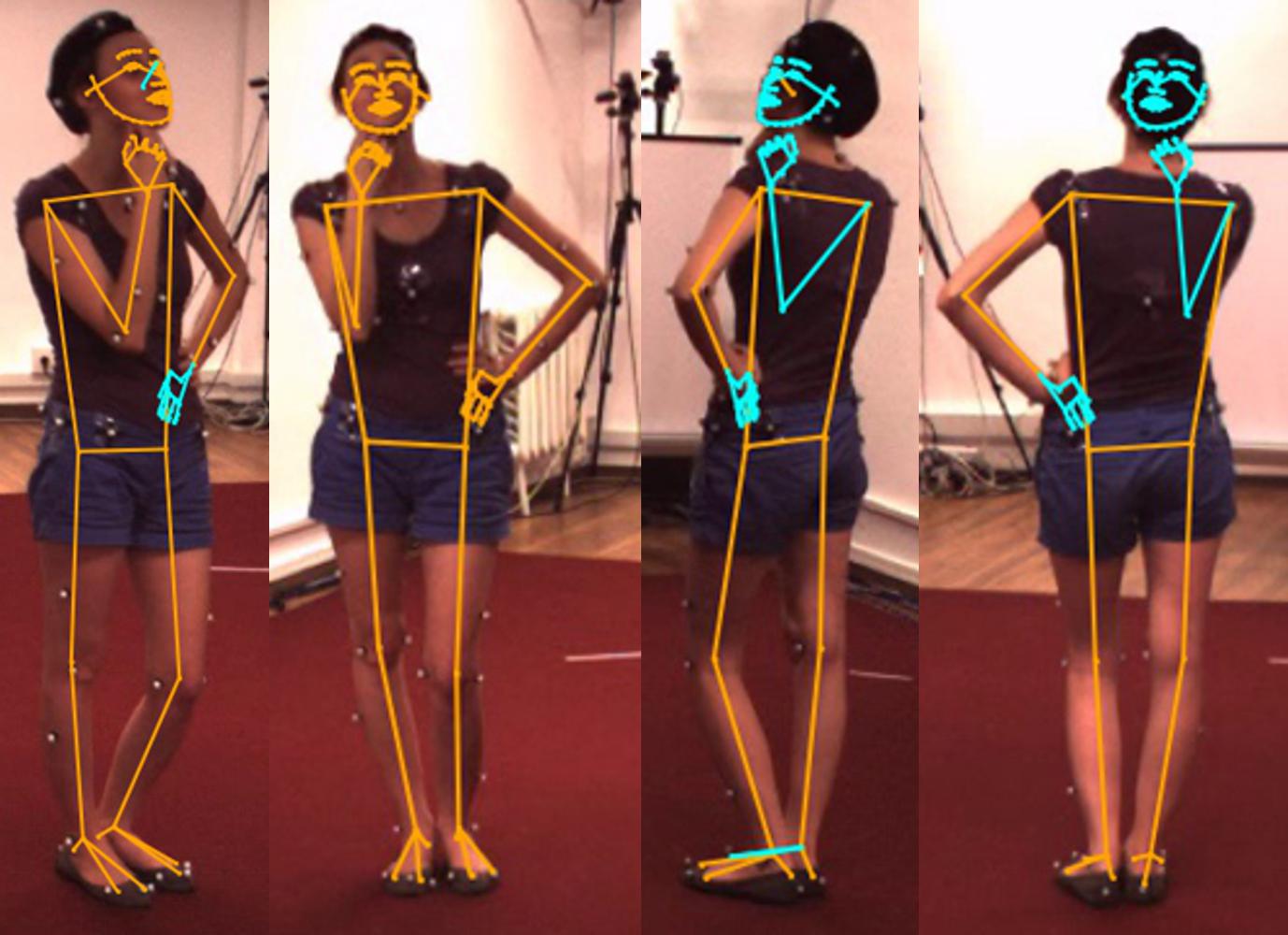}
  \caption{Example outputs of the completion network. The orange color denotes the keypoints that were detected by OpenPifPaf. The cyan color shows the missed keypoints by OpenPifPaf but completed by our completion network. The left hand is detected in only 1 view by OpenPifPaf and thus fully predicted by the completion network. }
  \label{fig:after-completion}
\end{figure}

\subsection{Completion network}
\label{sec:complete}

In order to complete the 26,333 incomplete 3D whole-body poses, we develop a completion network as shown in \autoref{fig:completionnet}. 
We designed our completion network using Transformer architecture~\cite{DBLP:journals/corr/VaswaniSPUJGKP17(transformer)} as they can easily handle the conditional dependencies introduced by the skeleton’s topology through masking. Since each skeleton always has exactly 133 keypoints, which can be considered as 133 tokens of 3 coordinate values. 
Token values are expanded from 3 coordinates to $3\times 16 = 48$ features using Fourier encoding.
We use learnable positional encoding since each keypoint is uniquely identified.

We train the completion network on the 11,426 complete skeletons using a masked auto-encoder strategy~\cite{he2022masked} where the missing keypoints are masked at the input and will be predicted using the unmasked keypoints.
The masking strategy is as follows:
\begin{itemize}\setlength\itemsep{0em}
    \item With a $50\%$ chance, we perform a keypoint wise mask where each keypoint has $15\%$ chance of being masked,
    \item with the remaining $50\%$ chance, we perform a block wise mask in which either the body, the left hand, the right hand, the left or the right part of the face are masked (uniform probability).
\end{itemize}

\begin{figure}[b]
  \centering
  \includegraphics[width=\linewidth]{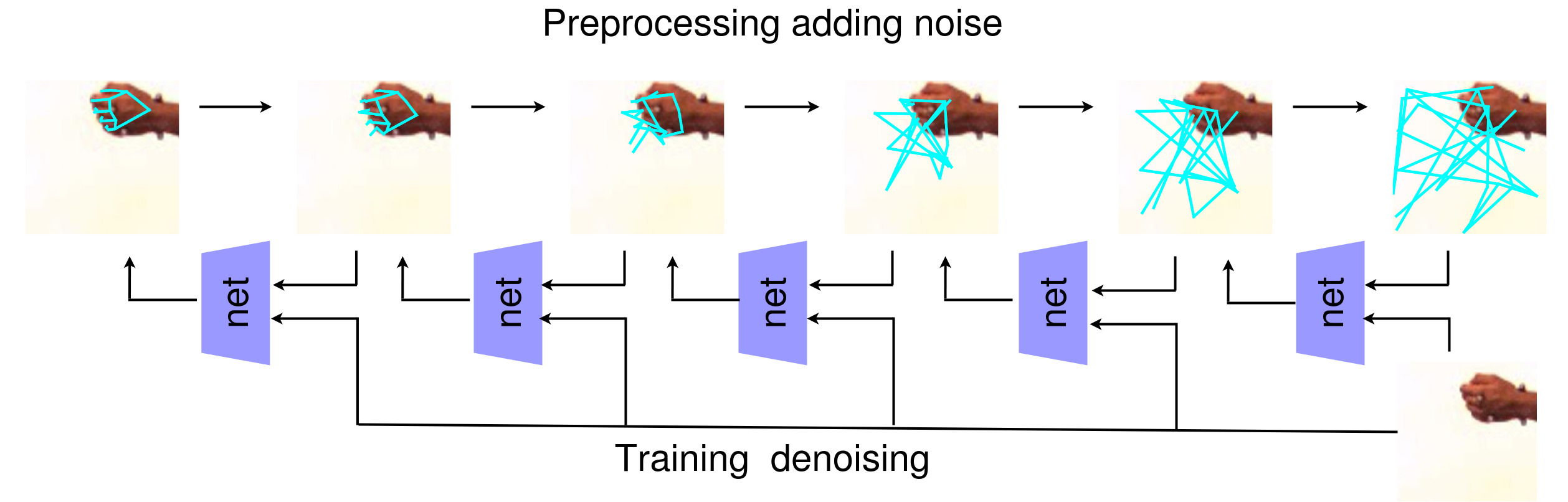}
  \caption{Refinement network architecture and training process. Gaussian noise is added to the groundtruth coordinates with increasing variance, and the network is iteratively trained to recover the less noisy coordinates.}
  \label{fig:BPN}
\end{figure}

\begin{figure}
  \centering
  \includegraphics[width=\linewidth]{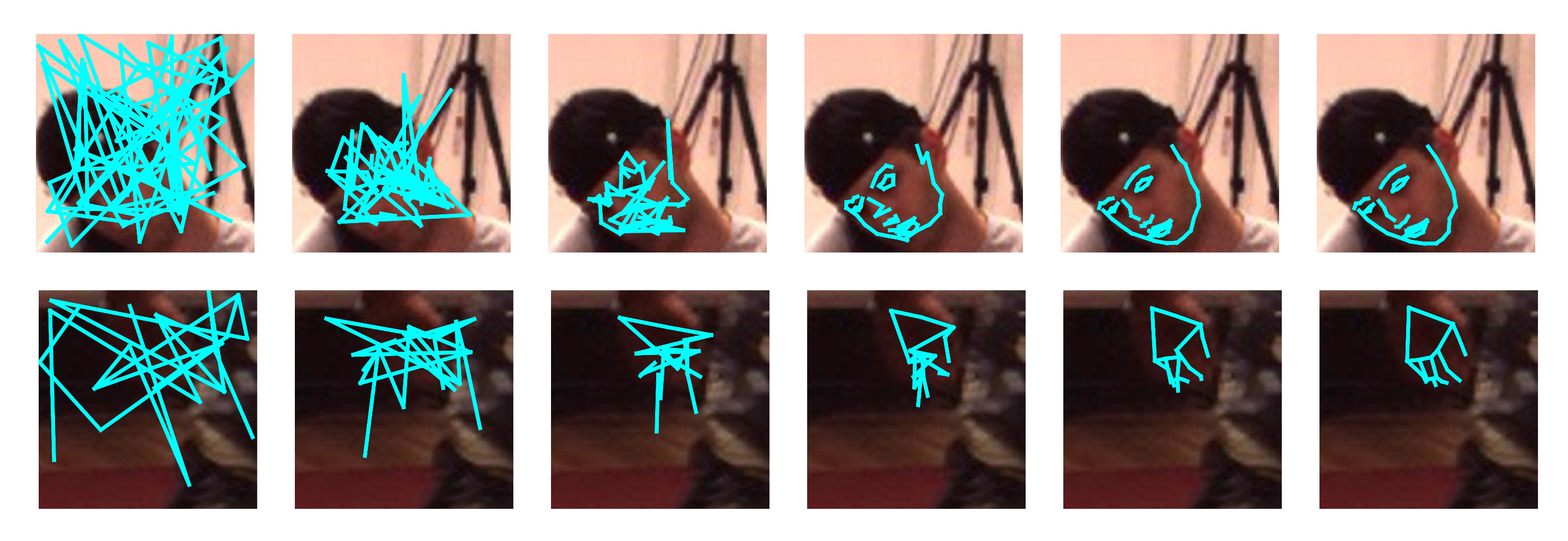}
  \caption{Example outputs from the face (top row) and hand (bottom row) refinement networks during inference time. We observe that the predictions almost converge to the correct locations in 5-iteration.}
  \label{fig:denoising_examples}
\end{figure}

To ease the learning process and take into account the causal link between some keypoints (\eg, the tip of a finger depends on the position of its parent phalanges), we introduce a curriculum approach.
We compute the loss at different levels following a hierarchy where early levels consider only keypoints closer to the root, while later levels consider more deformable keypoints which highly depend on their parents. We illustrate the completion network and learning process in \autoref{fig:completionnet}. 
The loss function is
\begin{align}
\nonumber \mathcal{L}(X, X_{gt3D}, X_{gt2D}) = & \mathcal{L}_{3D}(X, X_{gt3D}) \\
\nonumber    + & \alpha \mathcal{L}_{2D}(X, X_{gt2D})\\ 
    + & \beta \mathcal{L}_{sym}(X),
\end{align}
where $\mathcal{L}_3D$ is an $\ell_1$ loss of 3D coordinates, $\mathcal{L}_2D$ is an $\ell_1$ loss of 2D projection of the 3D coordinates if we have the 2D annotation from OpenPifPaf, and $\mathcal{L}_{sym}$ is a symmetric loss which is applied to make sure the left part and right part of the human have the same length on corresponding body parts.

\begin{figure*}
  \centering
  \includegraphics[width=\linewidth]{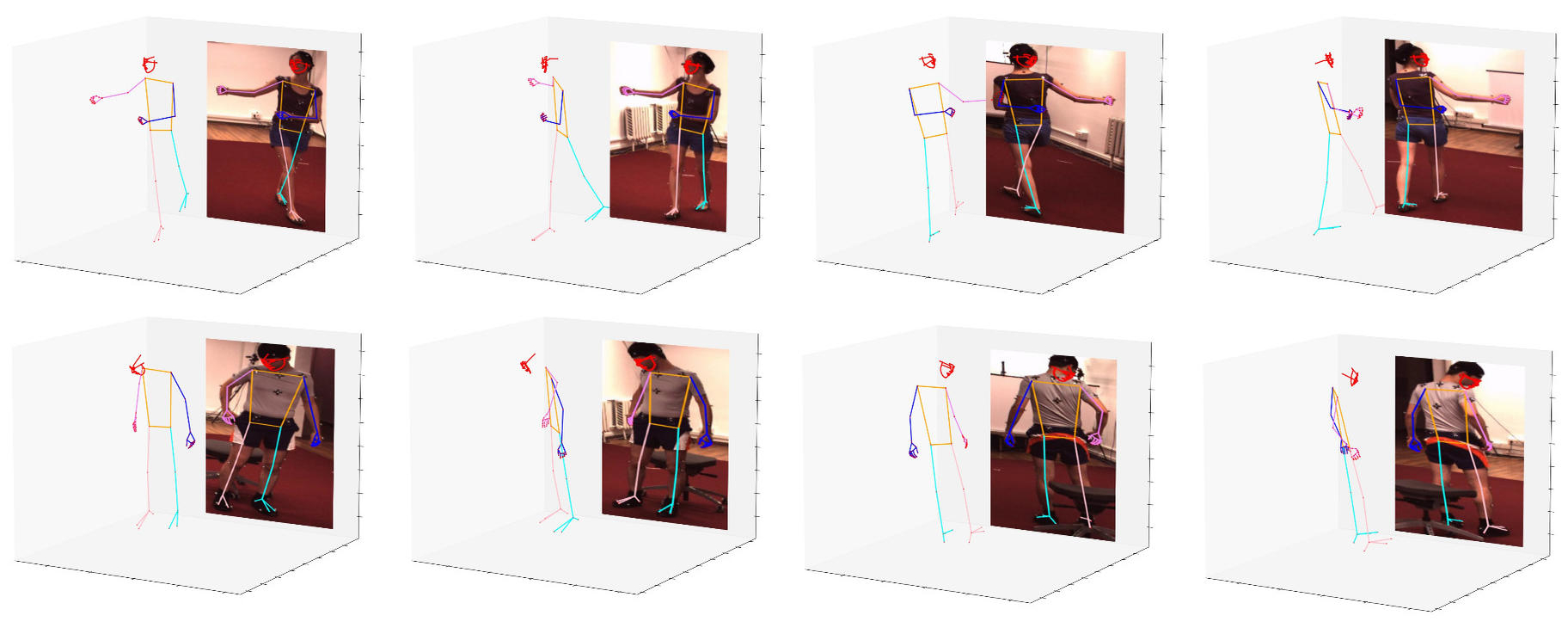}
  \caption{Examples of the 3D whole body skeleton. They are visually realistic humans. The strange looking faces (fatter or thinner) in different views are due to viewing artifacts of the default perspective projection.}
  \label{fig:benchmark_3d}
\end{figure*}

We show an example output from our completion network in \autoref{fig:after-completion}.
The completion network results on missing body parts are visually realistic and appealing.  
However, since the completion network does not rely on the image content, its output does not always align with the image and may only reflect the most common poses of the training set. This can quantitatively be seen in the line ``+ Completion'' of \autoref{tab:step_score}.

\subsection{Hands and face 2D refinements}

In order to correct the alignment problem, we propose another neural network that refines the 2D position of keypoints on the face and the hands. Previous studies have explored and demonstrated the effectiveness of 2D human pose refinement using an iterative error feedback framework~\cite{errorfeedback}. 
Motivated by this, we build upon recent conditional diffusion models \cite{DBLP:journals/corr/abs-2006-11239(diffusionmodel)} and we consider the prediction from the completion network as \emph{noisy} such that the refinement network \emph{denoises} it to conditionally fit the image.

We train separate refinement models for the face and the hands, while keeping the same network architecture and the same training strategy. We used a simple MLP and found it to be effective, preventing the need to explore more complex architectures. We illustrate the refinement process in \autoref{fig:BPN}.
During training, we add Gaussian noise to the groundtruth poses with an increasing variance from 5 to 25 pixels, and annotate them as step $t=1...5$ (step $t=0$ is the groundtruth). 
The network learns to predict the pose at step $t$ given the image and the noisier step $t+1$ with a 2D supervision loss.

We build two small datasets, each consisting of 22,000 non-occluded faces and hands respectively, with their corresponding OpenPifPaf predictions.
Each image is resized to $384 \times 384$ pixels. 
We use a random crop of size $224\times 224$ pixels to have the face and hands located in diverse regions of the images. 
We split the datasets into training and validation sets with 20,000 images and 2,000 images, respectively.

\begin{figure*}[t]
  \centering
  \includegraphics[width=0.81\linewidth]{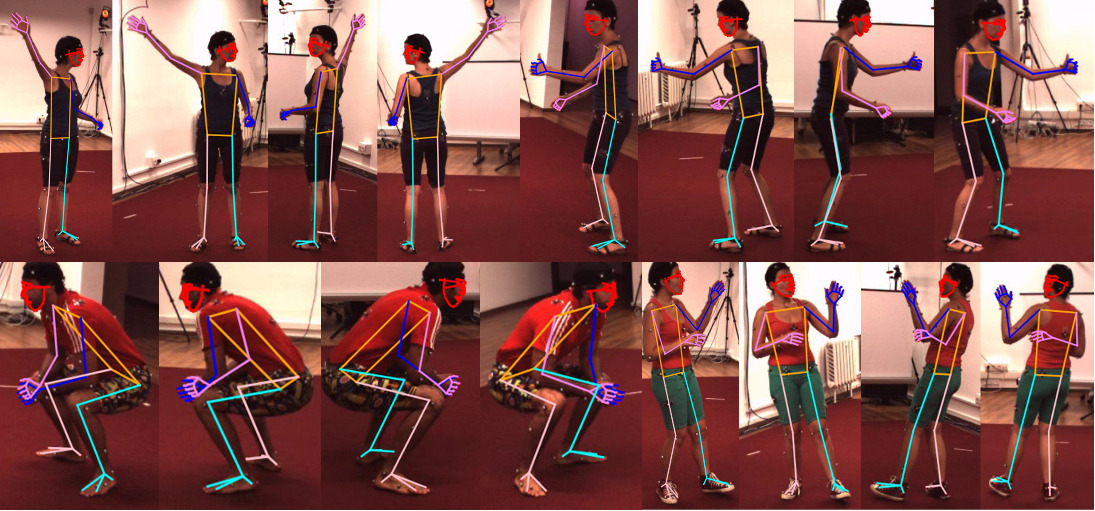}
  \caption{Examples of the 3D whole-body skeleton projected in 2D onto their corresponding images. They are visually accurate, though still there are small errors in detail which we do not expect to overcome due to the initial resolution and ambiguity of the images.}
  \label{fig:benchmark}
\end{figure*}

Quantitatively, the face predictions achieve an average error less than 3 pixels and the hand predictions achieve an average error less than 7 pixels on the validation sets.
We show example qualitative results in \autoref{fig:denoising_examples}.

Finally, we run the refinement networks on the 2D-projections of the 3D poses predicted by our completion network. 
For each 3D skeleton, we project it into the 4 different 2D views.
We then crop the regions around the hands and face and denoise the corresponding predictions using the refinement network with 10 iterations to obtain refined 2D poses in each of the 4 views. 

Although the refinement network is not always correct due to its training on non-occluded faces or hands, we only need 2 non-opposing views to perform geometric reconstruction. Since bad refinements tend to collapse all keypoints into the same location, we select the two non-opposing views with the highest variance in keypoint positions to avoid disruptions caused by occlusions. Using this method, we obtain 151,036 triplets of 3D whole-body keypoints, corresponding image, and 2D projected keypoints from the original set. Examples of resulting 3D whole-body skeletons and their image-aligned 2D counterparts are shown in \autoref{fig:benchmark_3d} and \autoref{fig:benchmark}, respectively.

\subsection{Quality assessment}

To select the most accurate triplets from our dataset, we reuse the refinement networks and employ a multi-crop strategy that accounts for the variance of the prediction. We project each 3D whole-body skeleton onto all 4 views, and produce four cropped images for each region of interest around the face and hands. The refinement network is run on these 4 crops, and the resulting predictions are aligned with the original prediction to compute the 2D error compared to the original 2D projection. We score the 3D skeletons by averaging the errors of all 4 projected views, and select the 5k lowest error skeletons from each subject of Human3.6M (S1, S5, S6, S7, S8) to form the $5\text{k}\times 4\text{(view)}\times 5\text{(subject)} = 100\text{k}$ triplets of $\{$image, 2D coordinates, 3D coordinates in camera space$\}$ of our 3D whole-body dataset.

\begin{figure}[h]
  \centering
  \includegraphics[width=0.48\linewidth]{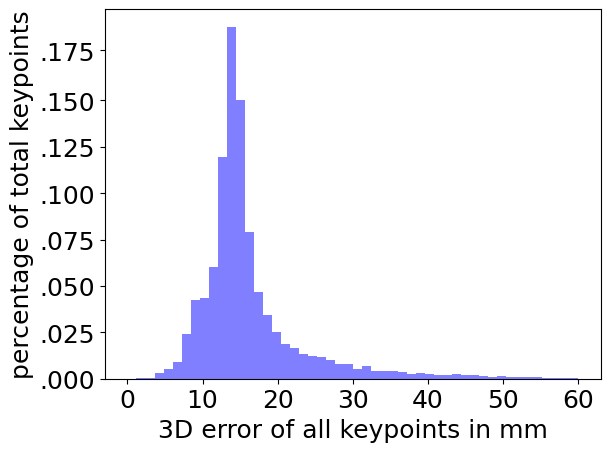} 
\includegraphics[width=0.48\linewidth]{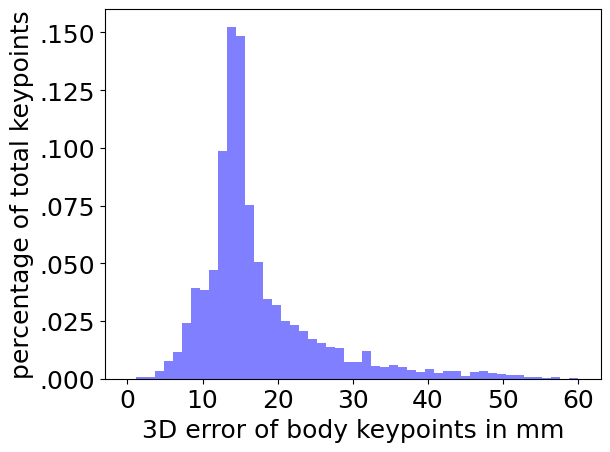} \\
\includegraphics[width=0.48\linewidth]{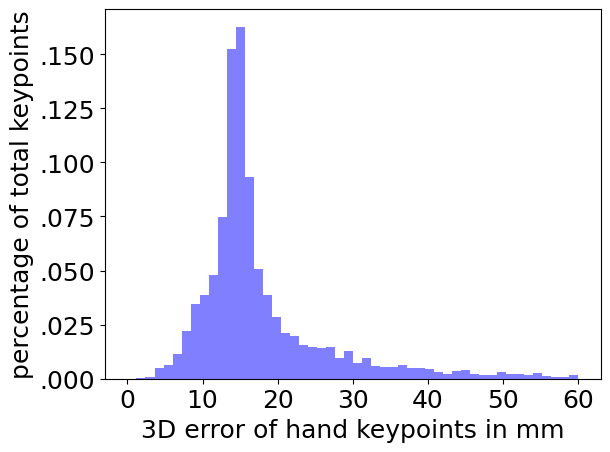}
\includegraphics[width=0.48\linewidth]{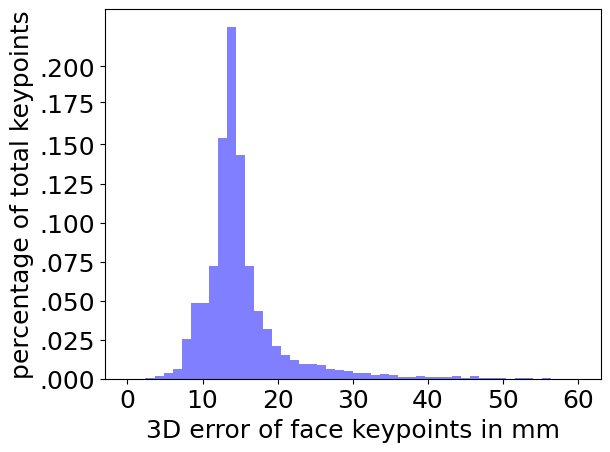}
  \caption{3D error distributions calculated from 80k manually corrected keypoint annotations. 3D error distributions are  presented for whole-body, body, hand and face in mm. We observe that 3D errors are mostly concentrated between 10mm and 20mm.}
  \label{fig:all_mm}
\end{figure}

\begin{figure}[h]
  \centering
  \includegraphics[width=0.97\linewidth]{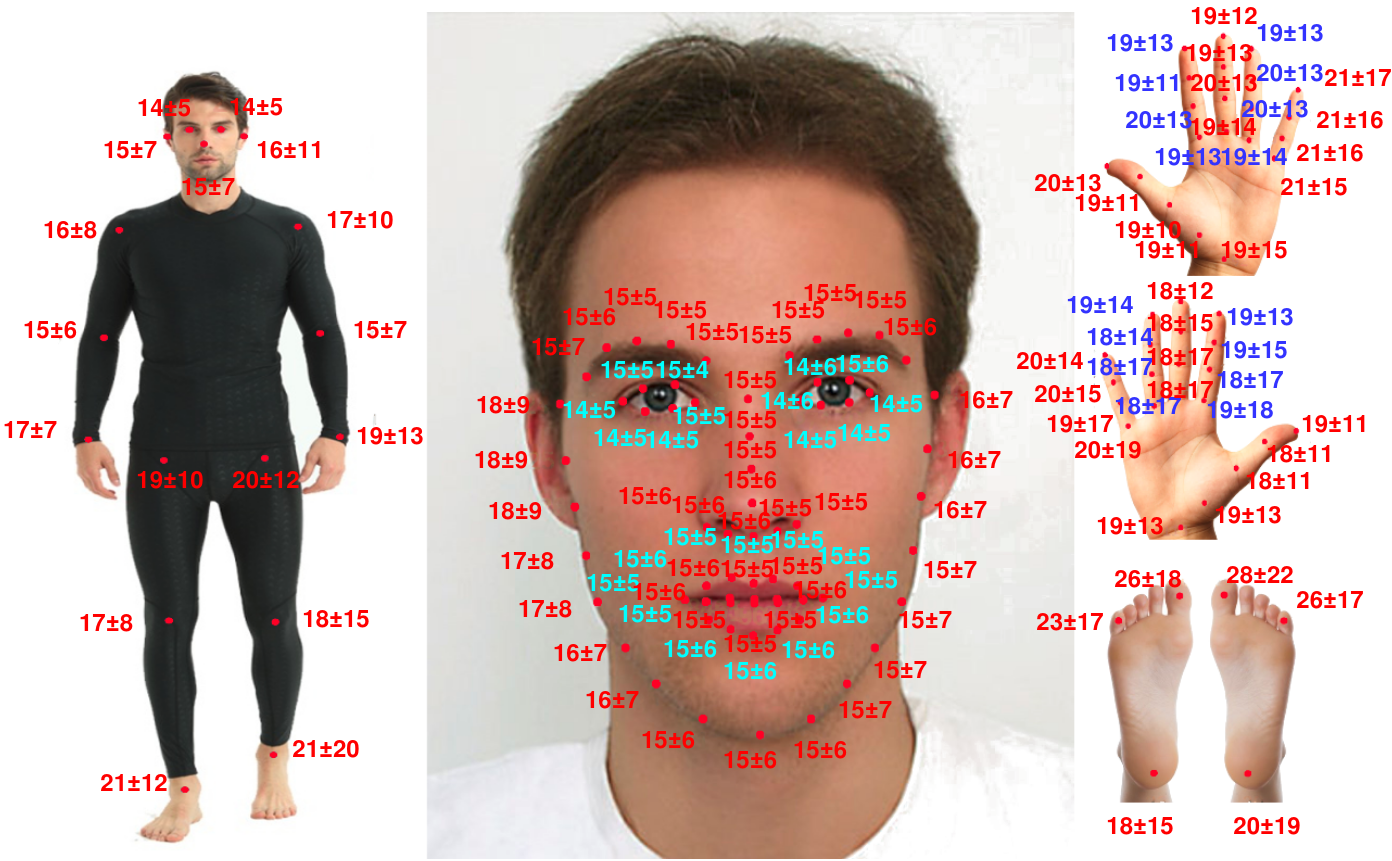}
  \caption{Per keypoints error statistics. Please zoom in.}
  \label{fig:90threshold}
\end{figure}

To assess the quality of the H3WB dataset, we conducted a cross-check study on 600 randomly selected images from the dataset.
In this study, annotators were presented an image with the 2D projection of the 3D skeleton on top and were asked to manually correct mis-aligned keypoints by drag and drop. Using multi-view geometry, we reconstructed these corrected skeletons in 3D and compared them to our original skeletons. 
To validate our process, we show the influence of each step in \autoref{tab:step_score}. The geometric approach produced good results but unfortunately cannot provide a large enough dataset. The completion step allows to obtain all labels but at the cost of degraded accuracy due to lack of alignment as explained in \autoref{sec:complete}. The diffusion recovers the original accuracy of the geometric approach. 2mm difference is irrelevant given the initial resolution of the images.
We obtain a final \textbf{average error of 17mm} which is very accurate for such a difficult task, and leads to a benchmark which we believe will not be saturated until methods reach around 35mm.

\begin{table}[h]
    \scriptsize
    \centering
    \begin{tabular}{lc |cccc }
     \toprule
\multicolumn{1}{l}{Steps} & \multicolumn{1}{c}{$\#$ keypoints available} &
    \multicolumn{4}{c}{3D error (mm)} \\
    \hline
    & & All & Body & Face & Hand \\
    Geometry & 48127 & 14.87 & 17.72 & 13.29 & 15.87 \\
    + Completion & 79800 & 29.31 & 25.57 & 26.02 & 36.67  \\
    + Diffusion & 79800 & 16.98 & 18.63 & 15.08 & 19.16  \\
    \bottomrule
    \end{tabular}
    \caption{Quantitative analysis of each intermediate step in our pipeline. }
    \label{tab:step_score}
\end{table}

\textbf{3D error distributions:} In \autoref{fig:all_mm}, we illustrate  3D error distributions in mm for all whole-body keypoints and each whole-body part keypoints (i.e. body, hand and face) separately. The distributions of errors are well concentrated around low values.

\textbf{Per keypoint errors:} All per keypoint error statistics are shown \autoref{fig:90threshold}. 98.3\% of the images are below 5cm error (97.4\% for body joints, 99.7\% hands, and 96.4\% face). Similarly, 82.6\% of the images are below 2cm error (75.5\% for body joints, 89.2\% hands, and 75.8\% face).

\subsection{TotalCapture 3D WholeBody Dataset}

In addition to H3WB dataset, we also prepare whole-body annotations for training sequence of poses fromTotalCapture~\cite{DBLP:journals/corr/abs-1801-01615(totalcapture)} dataset using our proposed multi-view pipeline. We call this dataset TotalCapture 3D WholeBody, or T3WB for short. To create TotalCapture 3D WholeBody, we first obtain 2D whole-body keypoints from OpenPifPaf~\cite{kreiss2021openpifpaf}. At this stage we discard the frames without any human. Next, we finetune the completion network initialized by H3WB weights using 7000 samples of TotalCapture with complete 8-views. At the end we obtain 125,960 triplets of 3D whole-body keypoint, corresponding image and 2D projected keypoints. This shows our pipeline can be used with any multiview dataset.

We do not conduct a quality assessment study for T3WB and therefore we cannot guarantee the precision of the annotations. Instead, we use T3WB together with H3WB for training models. To alleviate the effect of the noisy annotations in T3WB, we sample more from H3WB than T3WB in each batch. More specifically, we follow 4:1 ratio for each batch during \textit{H3WB + T3WB} trainings.

\section{The H3WB benchmark}

We use the H3WB dataset to propose a benchmark and the associated leaderboard. 
We split the dataset into training and test sets. The training set contains all samples from S1, S5, S6 and S7, including 80k \{image,2D,3D\} triplets. The test set contains all samples from S8, including 20k triplets.
The test set labels are retained to prevent involuntary overfitting on the test set. Evaluation is accessible only by submitting results to the maintainers. We do not provide a validation set. We encourage researchers to report 5-fold cross-validation average and standard deviation (see supplementary).

The corresponding benchmark has 3 different tasks:

\begin{enumerate}\setlength\itemsep{0em}
    \item 3D whole-body lifting from complete 2D whole-body skeletons, or \textit{2D$\rightarrow$3D} for short.
    \item 3D whole-body lifting from incomplete 2D whole-body skeletons, or \textit{I2D$\rightarrow$3D} for short.
    \item 3D whole-body skeleton prediction from image, or \textit{RGB$\rightarrow$3D} for short.
\end{enumerate}

For each task, we report the following MPJPE (Mean Per Joint Position Error) metrics:
\begin{itemize}\setlength\itemsep{0em}
    \item MPJPE for the whole-body, the body (keypoint 1-23), the face (keypoint 24-91) and the hands (keypoint 92-133) when whole-body is centered on the root joint, i.e. aligned with the pelvis, which in our case is the middle of two hip joints\footnote{We provide the whole-body  keypoint ids in supplementary material.},
    \item MPJPE for the face when it is centered on the nose, i.e. aligned with keypoint 1,
    \item MPJPE for the hands when hands are centered on the wrist, i.e left hand aligned with keypoint 92 and right hand aligned with keypoint 113.
\end{itemize}

To create baselines on each task, we adapt popular methods from the literature by changing the number of keypoints to that of our whole-body dataset. Notice that we keep the training recipes of the original papers to avoid over-fitting to this new benchmark.
In practice, we recommend to perform model selection and hyper-parameters tuning using 5-fold cross-validation.

\subsection{3D whole-body lifting from complete 2D whole-body keypoints (\textit{2D$\rightarrow$3D})} 
This task is similar to the standard 3D human pose estimation from 2D keypoints but using  whole-body keypoints. 
The training set contains 80k 2D-3D  pairs.
The test set contains only a half of all the test samples, \ie 10k 2D poses\footnote{The other half is reserved for the task I2D$\rightarrow$3D to prevent access to the missing keypoints.}.

We evaluate 6 methods on this task. SimpleBaseline~\cite{DBLP:journals/corr/MartinezHRL17} is a well-established model, consisting of a 6-layer MLP. We propose a modification, replacing the network architecture with an 8-layer MLP, which we call \textit{Large SimpleBaseline} inspired by CanonPose~\cite{DBLP:journals/corr/abs-2011-14679(canonpose)}. Jointformer~\cite{https://doi.org/10.48550/arxiv.2208.03704(jointformer)} is a recent transformer-based method. CanonPose is trained only with 2D supervision~\cite{DBLP:journals/corr/abs-2011-14679(canonpose)}. We also adapt CanonPose to work with additional 3D supervision by manually creating 3 fixed camera views and rotating the 3D skeletons into the corresponding view before projecting them into 2D, training it with multi-view weak-supervision. Finally, we report results for the parametric model SMPLify-X~\cite{smplifyx} by running optimizations on each input sample.

We train SimpleBaseline models using their official training setting as described in~\cite{DBLP:journals/corr/MartinezHRL17}. The inputs and targets are normalized by subtracting the mean and dividing by the standard deviation. Similarly, we train CanonPose~\cite{DBLP:journals/corr/abs-2011-14679(canonpose)} models following their official training setup where the inputs and targets are centered on the pelvis and scaled by the Forbenius norm.
We train the Jointformer model in the two stages as described in~\cite{https://doi.org/10.48550/arxiv.2208.03704(jointformer)}.

SimpleBaseline and CanonPose models output normalized whole-body keypoints which requires re-scaling at inference. 

We use statistics from the training set to adjust the test predictions. We calculate a scaling factor using the ratio of 3D to 2D bounding boxes. The formula is: $X_\text{final} = X_\text{unit} \times \overline{\sigma_{3d}} \times \frac{\sigma_{2d}}{\overline{\sigma_{2d}}}$, where $X_\text{unit}$ is the normalized prediction, $\overline{\sigma_{3d}}$ is the average size of the 3D training boxes, $\sigma_{2d}$ is the size of the current 2D box, and $\overline{\sigma_{2d}}$ is the average size of the 2D training boxes.

Since SMPLify-X has 144 keypoints with a different layout, we use interpolation to transform between the WholeBody skeleton and SMPL-X and run SMPL-X's optimization for 2,000 iterations (4 minutes/sample). 

\begin{table}[t]
    \centering
    \resizebox{\linewidth}{!}{
    \begin{tabular}{l c c c c}
     \toprule
    \multicolumn{1}{l}{Method} & \multicolumn{1}{c}{All} &
    \multicolumn{1}{c}{Body} &
    \multicolumn{1}{c}{Face / aligned$^\dagger$} &
    \multicolumn{1}{c}{Hand / aligned$^\ddagger$} \\
    \midrule
    \textit{H3WB} & & &  \\
    SMPL-X\cite{smplifyx} & 188.9 & 166.0 & 208.3 / 23.7 & 170.2 / 44.4 \\
    CanonPose\cite{DBLP:journals/corr/abs-2011-14679(canonpose)}$^*$ & 186.7 & 193.7 & 188.4 / 24.6 & 180.2 / 48.9 \\
    SimpleBaseline \cite{DBLP:journals/corr/MartinezHRL17}$^*$ & 125.4 & 125.7 & 115.9 / 24.6 & 140.7 / 42.5 \\
    CanonPose\cite{DBLP:journals/corr/abs-2011-14679(canonpose)} \textit{w} 3D sv.$^*$ & 117.7 & 117.5 & 112.0 / 17.9 & 126.9 / 38.3 \\
    Large SimpleBaseline\cite{DBLP:journals/corr/MartinezHRL17}$^*$ & 112.3 & 112.6 & 110.6 / \textbf{14.6} & \textbf{114.8} / \textbf{31.7} \\
    Jointformer\cite{https://doi.org/10.48550/arxiv.2208.03704(jointformer)} & \textbf{88.3} & \textbf{84.9} & \textbf{66.5} / 17.8 & 125.3 / 43.7 \\

    \midrule
    \textit{H3WB + T3WB} & & &  \\
    CanonPose\cite{DBLP:journals/corr/abs-2011-14679(canonpose)}$^*$ & 164.7 & 161.1 & 174.5 / 21.5 & 150.8 / 43.6 \\
    SimpleBaseline \cite{DBLP:journals/corr/MartinezHRL17}$^*$ & 115.3 & 114.8 & 109.4 / \textbf{15.8} & 125.1 / \textbf{33.5} \\
    Jointformer\cite{https://doi.org/10.48550/arxiv.2208.03704(jointformer)} & \textbf{81.5} & \textbf{78.0} & \textbf{60.4} / 16.2 & \textbf{117.6} / 38.8 \\
     \bottomrule
    \end{tabular}
    }
    \caption{Comparing different methods for 2D$\rightarrow$3D on H3WB test set. Results are shown for the MPJPE metric in mm. Methods with $^*$ output normalized predictions. Results of normalized methods are re-scaled using our scaling formula. All results are pelvis aligned, except $\dagger$ and $\ddagger$ show nose and wrist aligned results for face and hands, respectively. Sv. is supervision.}
    \label{tab:baseline_result1}
\end{table}

We present the results in \autoref{tab:baseline_result1}. SMPLify-X performs the worst, showing that parametric models struggle more than discriminative approaches. SimpleBaseline\cite{DBLP:journals/corr/MartinezHRL17} is a solid method, and Large SimpleBaseline improves its performance further. CanonPose\cite{DBLP:journals/corr/abs-2011-14679(canonpose)} can be improved with additional 3D supervision, but still performs worse than Large SimpleBaseline. CanonPose also predicts the camera view, and the uncertainty in this prediction can lead to more error. Jointformer\cite{https://doi.org/10.48550/arxiv.2208.03704(jointformer)} achieves the best results among all methods, but still has room for improvement. All methods perform worse on our benchmark than on Human3.6M because of pelvis centering, which creates higher numerical error on extremities like hands and face, the parts that contain most of the whole-body keypoints.

Additionally, we conducted experiments with SimpleBaseline, CanonPose, and Jointformer, leveraging the merged dataset \textit{H3WB + T3WB}, which combines both H3WB and T3WB. The results in \autoref{tab:baseline_result1} show that when T3WB is integrated with H3WB, the performances are improved significantly. For instance, on all whole-body keypoints, it yields 22 pt, 10.1 pt and 6.8 pt improvement for SimpleBaseline, CanonPose and Jointformer, respectively.

\subsection{3D whole-body lifting from incomplete 2D whole-body keypoints (\textit{I2D$\rightarrow$3D})} 
We propose a second task where we want to obtain 3D complete whole-body poses from 2D incomplete pose.
This task aims to simulate the more realistic case when there are occlusions and the 2D whole-body detector outputs an incomplete skeleton.
We do not provide masks for the training skeletons to allow for online data-augmentation.
Instead, we propose a masking strategy as follows:
\begin{itemize}\setlength\itemsep{0em}
    \item With 40\% probability, each keypoint has a 25\% chance of being masked,
    \item with 20\% probability, the face is entirely masked,
    \item with 20\% probability, the left hand is entirely masked,
    \item with 20\% probability, the right hand is entirely masked.
\end{itemize}
The second half of the test set (10k 2D) is devoted to this task. The masking strategy is applied only once on the 2D poses of the test set, which are directly provided as incomplete
2D skeletons for fair comparison between methods.

\begin{table}[b]
    \centering
    \resizebox{\linewidth}{!}{
    \begin{tabular}{lc c cc}
     \toprule
    \multicolumn{1}{l}{Method} & \multicolumn{1}{c}{All} &
    \multicolumn{1}{c}{Body} &
    \multicolumn{1}{c}{Face / aligned$^\dagger$} &
    \multicolumn{1}{c}{Hand / aligned$^\ddagger$} \\
    \hline
    CanonPose\cite{DBLP:journals/corr/abs-2011-14679(canonpose)}$^*$ & 285.0 & 264.4 & 319.7 / 31.9 & 240.0 / 56.2 \\
    SimpleBaseline\cite{DBLP:journals/corr/MartinezHRL17}$^*$ & 268.8 & 252.0 & 227.9 / 34.0 & 344.3 / 83.4 \\
    CanonPose\cite{DBLP:journals/corr/abs-2011-14679(canonpose)} + 3D sv.$^*$ & 163.6 & 155.9 & 161.3 / 22.2 & 171.4 / 47.4 \\
    Large SimpleBaseline\cite{DBLP:journals/corr/MartinezHRL17}$^*$ & 131.4 & 131.6 & 120.6 / \textbf{19.8} & \textbf{148.8} / \textbf{44.8} \\
    Jointformer\cite{https://doi.org/10.48550/arxiv.2208.03704(jointformer)} & \textbf{109.2} & \textbf{103.0} & \textbf{82.4} / \textbf{19.8} & 155.9 / 53.5 \\
    \midrule
    \textit{H3WB + T3WB} & & &  \\
    CanonPose\cite{DBLP:journals/corr/abs-2011-14679(canonpose)}$^*$ & 261.5 & 243.3 & 291.3 / 31.3 & 223.1 / 53.7 \\
    SimpleBaseline\cite{DBLP:journals/corr/MartinezHRL17}$^*$ & 260.5 & 238.0 & 221.1 / 32.2 & 336.5 / 80.4 \\
    Jointformer\cite{https://doi.org/10.48550/arxiv.2208.03704(jointformer)} & \textbf{84.2} & \textbf{80.1} & \textbf{59.4} / \textbf{16.3} & \textbf{126.5}/ \textbf{44.5} \\
    \bottomrule
    \end{tabular}}
    \caption{Comparing different methods for I2D$\rightarrow$3D on H3WB test set. Results are shown for the MPJPE metric in mm. Methods with $^*$ output normalized predictions. Results of normalized methods are re-scaled using our scaling formula. All results are pelvis aligned, except $\dagger$ and $\ddagger$ show nose and wrist aligned results for face and hands, respectively. Sv. is supervision.}
    \label{tab:baseline_result2}
\end{table}

\begin{figure}[t]
  \centering
  \includegraphics[width=0.495\linewidth]{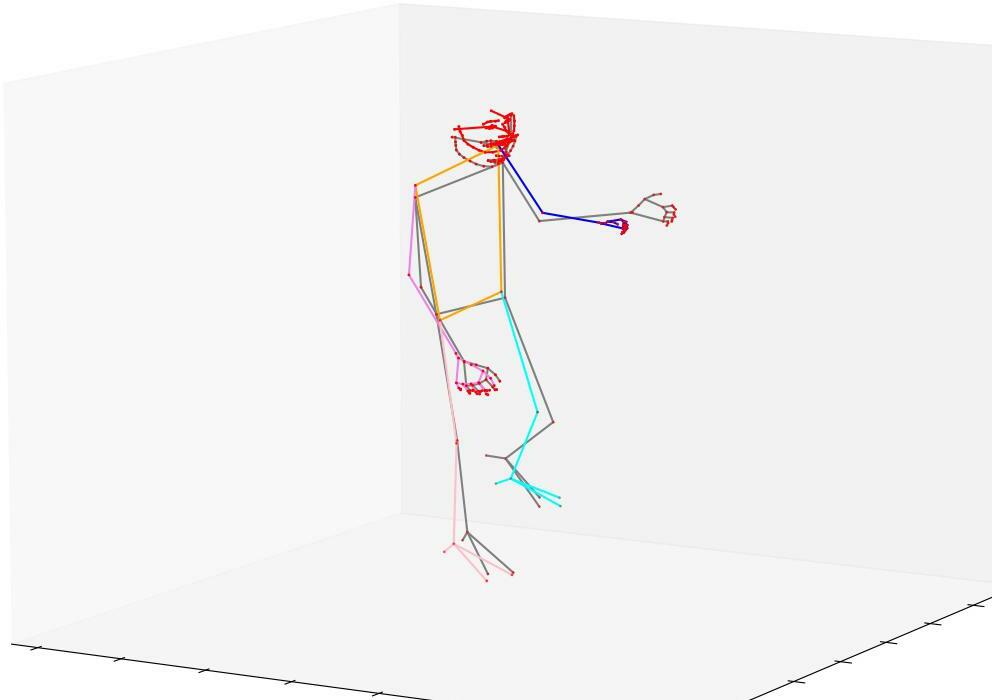}\includegraphics[width=0.495\linewidth]{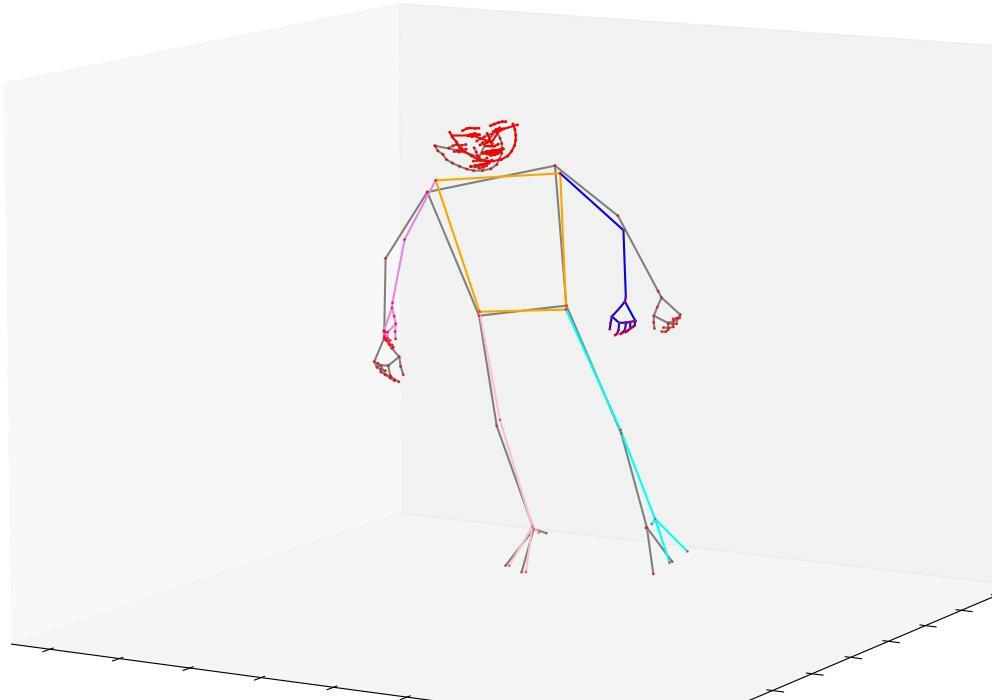}
  \includegraphics[width=0.495\linewidth]{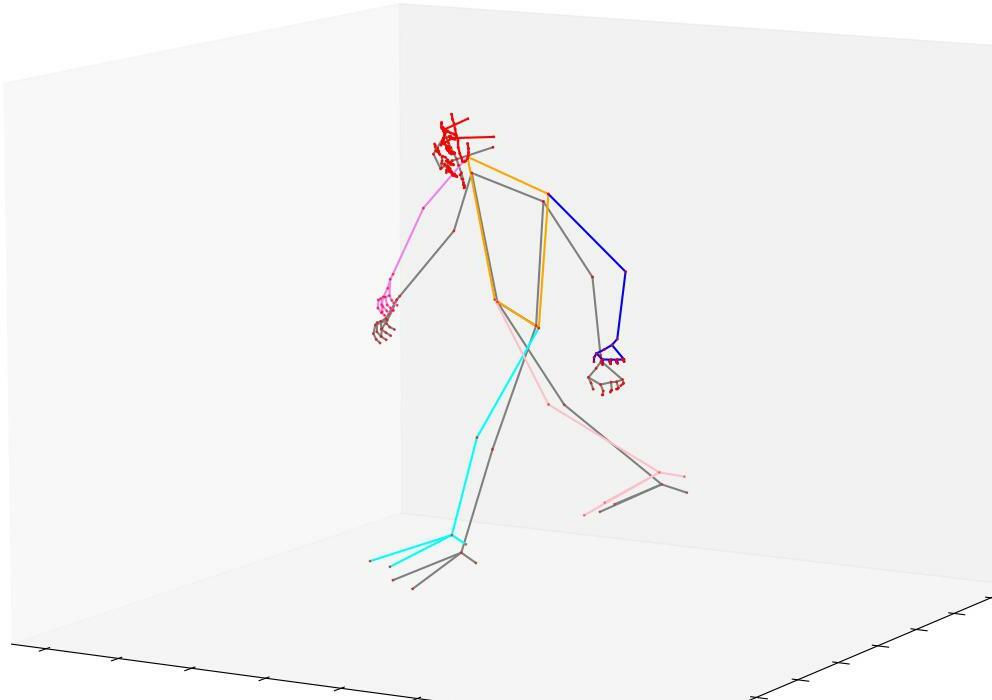}\includegraphics[width=0.495\linewidth]{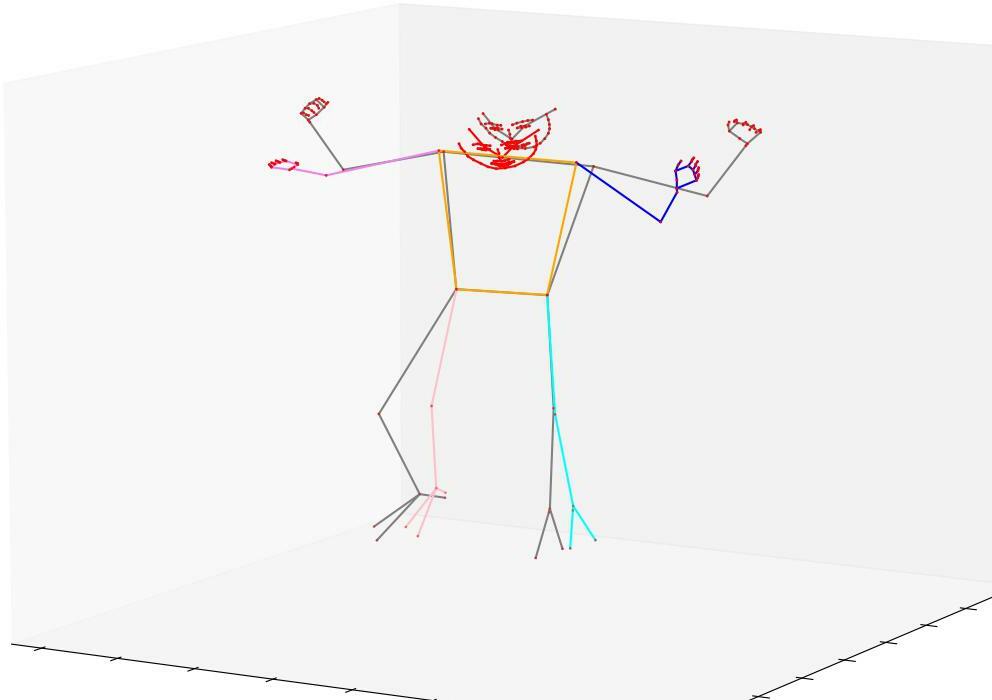}
  \caption{Example predictions from Large SimpleBaseline model for task  I2D$\rightarrow$3D. Colored skeletons correspond to predictions and gray skeletons correspond to  groundtruths.}
  \label{fig:task2}
\end{figure}

The results for the I2D$\rightarrow$3D task are shown in \autoref{tab:baseline_result2}. All methods perform worse than in the 2D$\rightarrow$3D task. SimpleBaseline\cite{DBLP:journals/corr/MartinezHRL17} has low capacity and uses batch normalization that struggles with missing data, resulting in poor performance. The Large SimpleBaseline model, without batch normalization layers, achieves good results for the task's complexity. CanonPose\cite{DBLP:journals/corr/abs-2011-14679(canonpose)} performs poorly due to errors in camera rotation prediction, which are magnified since most of the 133 keypoints are on the face and hands. The addition of 3D supervision partly solves this problem. The transformer-based Jointformer\cite{https://doi.org/10.48550/arxiv.2208.03704(jointformer)} method outperforms others. Sample outputs obtained by Large SimpleBaseline are shown in \autoref{fig:task2}, where predicted skeletons, although not accurate, are realistic.

Similar to 2D$\rightarrow$3D task, we experimented with SimpleBaseline, CanonPose and Jointformer on \textit{H3WB + T3WB} dataset and obtained significant improvements. For  all whole-body keypoints, the combined dataset yields 23.5 pt, 8.3 pt and 25 pt improvement for SimpleBaseline, CanonPose and Jointformer, respectively. Those results further validate the effectiveness of our multi-view pipeline for creating whole-body keypoint annotations for any multi-view dataset. 

\subsection{3D whole-body pose estimation from a single image (\textit{RGB$\rightarrow$3D})} 
This task is the standard monocular 3D human pose estimation task extended to whole-body pose estimation. 
We provide a script to split the original Human3.6M videos into images with our indexing in order to establish image-3D correspondences. 
The training set contains 80k \{image paths,3D\} pairs, as well as the 2D bounding box of the human in the image. 
The test set contains all the test samples, including 20k image paths and their 2D bounding boxes. 
2D coordinates are not given in order to avoid collisions with 2D$\rightarrow$3D and I2D$\rightarrow$3D.

For this task, we train 2 two-stage models and 1 single-stage model. For our first two-stage model, we first train a Stacked Hourglass Network (SHN)\cite{DBLP:journals/corr/NewellYD16_(stackhourglass)} to predict 2D whole-body keypoints. Then, SimpleBaseline\cite{DBLP:journals/corr/MartinezHRL17} takes 2D keypoint predictions as input and lifts them to 3D space. Similarly, the second two-stage model utilizes CPN\cite{DBLP:journals/corr/abs-1711-07319(CPN)} to output 2D keypoints and then Jointformer\cite{https://doi.org/10.48550/arxiv.2208.03704(jointformer)} lifts the 2D predictions to obtain 3D whole-body poses. For our single-stage model, we modify the last layer of Resnet50\cite{DBLP:journals/corr/HeZRS15(Resnet)} to directly output the 3D whole-body keypoints. We regress the 3D whole-body keypoint coordinates using L1 loss.

\begin{table}[th]
    \centering
    \resizebox{\linewidth}{!}{
    \begin{tabular}{l c  c  c c}
      \toprule
    \multicolumn{1}{l}{Method} & \multicolumn{1}{c}{All} &
    \multicolumn{1}{c}{Body} &
    \multicolumn{1}{c}{Face / aligned $^\dagger$} &
    \multicolumn{1}{c}{Hand / aligned$^\ddagger$} \\
    \midrule
    \textbf{RGB$\rightarrow$2D+2D$\rightarrow$3D:} & & & & \\
    \multicolumn{1}{r}{~~SHN\cite{DBLP:journals/corr/NewellYD16_(stackhourglass)}+SimpleBaseline$^*$} & 182.5 & 189.6 & 138.7 / 32.5 & 249.4 / 64.3 \\
    \multicolumn{1}{r}{CPN\cite{DBLP:journals/corr/abs-1711-07319(CPN)}+Jointformer\cite{https://doi.org/10.48550/arxiv.2208.03704(jointformer)}} & \textbf{132.6} & \textbf{142.8} & \textbf{91.9} / \textbf{20.7} & \textbf{192.7} / \textbf{56.9} \\
    \textbf{RGB$\rightarrow$3D:} & & & & \\
    \multicolumn{1}{r}{Resnet50\cite{DBLP:journals/corr/HeZRS15(Resnet)}} & 166.7 & 151.6 & 123.6 / 26.3 & 244.9 / 63.1 \\
     \multicolumn{1}{r}{DOPE\cite{dope}} & 191.3 & 199.7 & 187.3 / 66.0 & 193.3 / 78.2 \\
     \bottomrule
    \end{tabular}}
    \caption{Comparing different methods for RGB$\rightarrow$3D on H3WB test set. Results are shown for the MPJPE metric in mm. Methods with $^*$ output normalized predictions. Results of normalized methods are re-scaled using our scaling formula. All results are pelvis aligned, except $\dagger$ and $\ddagger$ show nose and wrist aligned results for face and hands, respectively.  }
    \label{tab:baseline_result3}
\end{table}

Results in \autoref{tab:baseline_result3} show the two-stage \textit{CPN + Jointformer} model obtains the best results. Our simple single-stage method performs better than the two-stage \textit{SHN + SimpleBaseline} model. Learning 2D whole-body keypoints is challenging for SHN as very close keypoints on face and hands may introduce noise to the predicted keypoint heatmaps. The error in the 2D keypoints then makes the lifting task much more challenging. 
Surprisingly, RGB$\rightarrow$3D seems to be harder that the I2D$\rightarrow$3D task. Although there are also missing body parts due to self occlusion, RGB$\rightarrow$3D contains more contextual information that should allow to better disambiguate the pose.
Compared to 2D$\rightarrow$3D and I2D$\rightarrow$3D, direct prediction of 3D whole-body pose from images remains thus as a challenging task which we hope this benchmark can help improve over time.

In order to show the importance of training body parts jointly, we evaluate DOPE~\cite{dope} on our benchmark. Unfortunately, it fails to address occluded body parts only predicts the whole-body keypoints for 35\% of the test set. For each missing keypoint, we use the (topological) nearest predicted joint as a proxy. Even so, a disjointed model like DOPE fails to achieve significant accuracy.

\section{Conclusion}

In this paper, we introduce the H3WB dataset, which extends the Human3.6M dataset with 2D and 3D keypoint annotations for body, face, and hands, containing 100k images with 133 keypoints with an average accuracy of 17mm. We propose three tasks based on this dataset: 3D whole-body lifting from complete 2D keypoints, 3D whole-body lifting from incomplete 2D keypoints, and 3D whole-body prediction from monocular images. We evaluate several baselines on these tasks and demonstrate promising accuracy, but with room for improvement. Lifting from incomplete 2D skeletons and direct estimation from monocular images remain challenging, and we hope that our dataset and benchmark will spur future research in these areas.

\section*{Acknowledgments}
This work was supported by ANR project TOSAI ANR-20-IADJ-
0009 and Ergonova Conseil, and was granted access to the HPC resources of IDRIS under the allocation 2023-AD011012640R2 and 2023-AD011013267R1 made by GENCI. 

\clearpage
{\small
\bibliographystyle{ieee_fullname}
\bibliography{egbib}

\begin{thebibliography}{10}\itemsep=-1pt

\bibitem{arnab2019exploiting}
Anurag Arnab, Carl Doersch, and Andrew Zisserman.
\newblock Exploiting temporal context for 3d human pose estimation in the wild.
\newblock In {\em CVPR}, 2019.

\bibitem{baek2019pushing}
Seungryul Baek, Kwang~In Kim, and Tae-Kyun Kim.
\newblock Pushing the envelope for rgb-based dense 3d hand pose estimation via
  neural rendering.
\newblock In {\em CVPR}, 2019.

\bibitem{bagautdinov2021driving}
Timur Bagautdinov, Chenglei Wu, Tomas Simon, Fabian Prada, Takaaki Shiratori,
  Shih-En Wei, Weipeng Xu, Yaser Sheikh, and Jason Saragih.
\newblock Driving-signal aware full-body avatars.
\newblock {\em TOG}, 2021.

\bibitem{article(completion2)}
Dennis Bautembach, Iason Oikonomidis, and Antonis Argyros.
\newblock Filling the joints: Completion and recovery of incomplete 3d human
  poses.
\newblock {\em Technologies}, 2018.

\bibitem{blanz1999morphable}
Volker Blanz and Thomas Vetter.
\newblock A morphable model for the synthesis of 3d faces.
\newblock In {\em Proceedings of the 26th annual conference on Computer
  graphics and interactive techniques}, 1999.

\bibitem{DBLP:journals/corr/BogoKLG0B16(SMPLifyModel)}
Federica Bogo, Angjoo Kanazawa, Christoph Lassner, Peter~V. Gehler, Javier
  Romero, and Michael~J. Black.
\newblock Keep it {SMPL:} automatic estimation of 3d human pose and shape from
  a single image.
\newblock {\em ECCV}, 2016.

\bibitem{boukhayma20193d}
Adnane Boukhayma, Rodrigo~de Bem, and Philip~HS Torr.
\newblock 3d hand shape and pose from images in the wild.
\newblock In {\em CVPR}, 2019.

\bibitem{cai2018weakly}
Yujun Cai, Liuhao Ge, Jianfei Cai, and Junsong Yuan.
\newblock Weakly-supervised 3d hand pose estimation from monocular rgb images.
\newblock In {\em ECCV}, 2018.

\bibitem{cao2018pose}
Kaidi Cao, Yu Rong, Cheng Li, Xiaoou Tang, and Chen~Change Loy.
\newblock Pose-robust face recognition via deep residual equivariant mapping.
\newblock In {\em CVPR}, 2018.

\bibitem{DBLP:conf/eccv/CarissimiRBM18(completion1)}
Nicolò Carissimi, Paolo Rota, Cigdem Beyan, and Vittorio Murino.
\newblock Filling the gaps: Predicting missing joints of human poses using
  denoising autoencoders.
\newblock In {\em ECCV Workshops}, 2018.

\bibitem{errorfeedback}
Joao Carreira, Pulkit Agrawal, Katerina Fragkiadaki, and Jitendra Malik.
\newblock Human pose estimation with iterative error feedback.
\newblock In {\em Proceedings of the IEEE conference on computer vision and
  pattern recognition}, pages 4733--4742, 2016.

\bibitem{IonescuSminchisescu11}
Cristian~Sminchisescu Catalin~Ionescu, Fuxin~Li.
\newblock Latent structured models for human pose estimation.
\newblock In {\em ICCV}, 2011.

\bibitem{chang2018expnet}
Feng-Ju Chang, Anh~Tuan Tran, Tal Hassner, Iacopo Masi, Ram Nevatia, and Gerard
  Medioni.
\newblock Expnet: Landmark-free, deep, 3d facial expressions.
\newblock In {\em FG 2018}, 2018.

\bibitem{chen2021camera}
Xingyu Chen, Yufeng Liu, Chongyang Ma, Jianlong Chang, Huayan Wang, Tian Chen,
  Xiaoyan Guo, Pengfei Wan, and Wen Zheng.
\newblock Camera-space hand mesh recovery via semantic aggregation and adaptive
  2d-1d registration.
\newblock In {\em CVPR}, 2021.

\bibitem{chen2021model}
Yujin Chen, Zhigang Tu, Di Kang, Linchao Bao, Ying Zhang, Xuefei Zhe, Ruizhi
  Chen, and Junsong Yuan.
\newblock Model-based 3d hand reconstruction via self-supervised learning.
\newblock In {\em CVPR}, 2021.

\bibitem{DBLP:journals/corr/abs-1711-07319(CPN)}
Yilun Chen, Zhicheng Wang, Yuxiang Peng, Zhiqiang Zhang, Gang Yu, and Jian Sun.
\newblock Cascaded pyramid network for multi-person pose estimation.
\newblock {\em CVPR}, 2017.

\bibitem{Choi_2021_CVPR(TempConsistency}
Hongsuk Choi, Gyeongsik Moon, Ju~Yong Chang, and Kyoung~Mu Lee.
\newblock Beyond static features for temporally consistent 3d human pose and
  shape from a video.
\newblock In {\em CVPR}, 2021.

\bibitem{expose}
Vasileios Choutas, Georgios Pavlakos, Timo Bolkart, Dimitrios Tzionas, and
  Michael~J Black.
\newblock Monocular expressive body regression through body-driven attention.
\newblock In {\em ECCV}, 2020.

\bibitem{crispell2017pix2face}
Daniel Crispell and Maxim Bazik.
\newblock Pix2face: Direct 3d face model estimation.
\newblock In {\em ICCV Workshops}, pages 2512--2518, 2017.

\bibitem{DBLP:journals/corr/abs-1903-08527(face2)}
Yu Deng, Jiaolong Yang, Sicheng Xu, Dong Chen, Yunde Jia, and Xin Tong.
\newblock Accurate 3d face reconstruction with weakly-supervised learning: From
  single image to image set.
\newblock {\em CVPR Workshop}, 2019.

\bibitem{feng2018joint}
Yao Feng, Fan Wu, Xiaohu Shao, Yanfeng Wang, and Xi Zhou.
\newblock Joint 3d face reconstruction and dense alignment with position map
  regression network.
\newblock In {\em ECCV}, pages 534--551, 2018.

\bibitem{fieraru2020three}
Mihai Fieraru, Mihai Zanfir, Elisabeta Oneata, Alin-Ionut Popa, Vlad Olaru, and
  Cristian Sminchisescu.
\newblock Three-dimensional reconstruction of human interactions.
\newblock In {\em Proceedings of the IEEE/CVF Conference on Computer Vision and
  Pattern Recognition}, pages 7214--7223, 2020.

\bibitem{fieraru2021learning}
Mihai Fieraru, Mihai Zanfir, Elisabeta Oneata, Alin-Ionut Popa, Vlad Olaru, and
  Cristian Sminchisescu.
\newblock Learning complex 3d human self-contact.
\newblock In {\em Proceedings of the AAAI Conference on Artificial
  Intelligence}, volume~35, pages 1343--1351, 2021.

\bibitem{afit}
Mihai Fieraru, Mihai Zanfir, Silviu~Cristian Pirlea, Vlad Olaru, and Cristian
  Sminchisescu.
\newblock Aifit: Automatic 3d human-interpretable feedback models for fitness
  training.
\newblock In {\em Proceedings of the IEEE/CVF Conference on Computer Vision and
  Pattern Recognition}, pages 9919--9928, 2021.

\bibitem{garcia2019human}
Mercedes Garcia-Salguero, Javier Gonzalez-Jimenez, and Francisco-Angel Moreno.
\newblock Human 3d pose estimation with a tilting camera for social mobile
  robot interaction.
\newblock {\em Sensors}, 2019.

\bibitem{DBLP:journals/corr/abs-2001-02024}
Erik G{\"{a}}rtner, Aleksis Pirinen, and Cristian Sminchisescu.
\newblock Deep reinforcement learning for active human pose estimation.
\newblock {\em AAAI}, 2020.

\bibitem{ge20193d}
Liuhao Ge, Zhou Ren, Yuncheng Li, Zehao Xue, Yingying Wang, Jianfei Cai, and
  Junsong Yuan.
\newblock 3d hand shape and pose estimation from a single rgb image.
\newblock In {\em CVPR}, pages 10833--10842, 2019.

\bibitem{gu2019home}
Yiwen Gu, Shreya Pandit, Elham Saraee, Timothy Nordahl, Terry Ellis, and
  Margrit Betke.
\newblock Home-based physical therapy with an interactive computer vision
  system.
\newblock In {\em Proceedings of the IEEE/CVF International Conference on
  Computer Vision Workshops}, 2019.

\bibitem{gui2018teaching}
Liang-Yan Gui, Kevin Zhang, Yu-Xiong Wang, Xiaodan Liang, Jos{\'e}~MF Moura,
  and Manuela Veloso.
\newblock Teaching robots to predict human motion.
\newblock In {\em IROS}, 2018.

\bibitem{habibie2019wild}
Ikhsanul Habibie, Weipeng Xu, Dushyant Mehta, Gerard Pons-Moll, and Christian
  Theobalt.
\newblock In the wild human pose estimation using explicit 2d features and
  intermediate 3d representations.
\newblock {\em CVPR}, 2019.

\bibitem{hampali2021handsformer}
Shreyas Hampali, Sayan~Deb Sarkar, Mahdi Rad, and Vincent Lepetit.
\newblock Handsformer: Keypoint transformer for monocular 3d pose estimation
  ofhands and object in interaction.
\newblock {\em CVPR}, 2022.

\bibitem{he2022masked}
Kaiming He, Xinlei Chen, Saining Xie, Yanghao Li, Piotr Doll{\'a}r, and Ross
  Girshick.
\newblock Masked autoencoders are scalable vision learners.
\newblock In {\em CVPR}, 2022.

\bibitem{DBLP:journals/corr/HeZRS15(Resnet)}
Kaiming He, Xiangyu Zhang, Shaoqing Ren, and Jian Sun.
\newblock Deep residual learning for image recognition.
\newblock {\em CVPR}, 2015.

\bibitem{DBLP:journals/corr/abs-2005-04551(hand2)}
Yihui He, Rui Yan, Katerina Fragkiadaki, and Shoou{-}I Yu.
\newblock Epipolar transformers.
\newblock {\em CVPR}, 2020.

\bibitem{DBLP:journals/corr/abs-2006-11239(diffusionmodel)}
Jonathan Ho, Ajay Jain, and Pieter Abbeel.
\newblock Denoising diffusion probabilistic models.
\newblock {\em NeurIPS}, 2020.

\bibitem{journals/pami/IonescuPOS14(H36m)}
Catalin Ionescu, Dragos Papava, Vlad Olaru, and Cristian Sminchisescu.
\newblock Human3.6m: Large scale datasets and predictive methods for 3d human
  sensing in natural environments.
\newblock {\em TPMAI}, 2014.

\bibitem{iqbal2018hand}
Umar Iqbal, Pavlo Molchanov, Thomas Breuel~Juergen Gall, and Jan Kautz.
\newblock Hand pose estimation via latent 2.5 d heatmap regression.
\newblock In {\em ECCV}, 2018.

\bibitem{Iqbal_2020_CVPR(MultiviewInTheWild)}
Umar Iqbal, Pavlo Molchanov, and Jan Kautz.
\newblock Weakly-supervised 3d human pose learning via multi-view images in the
  wild.
\newblock In {\em CVPR}, 2020.

\bibitem{jackson2017large}
Aaron~S Jackson, Adrian Bulat, Vasileios Argyriou, and Georgios Tzimiropoulos.
\newblock Large pose 3d face reconstruction from a single image via direct
  volumetric cnn regression.
\newblock In {\em ICCV}, 2017.

\bibitem{jin2020whole(COCO-WholeBody)}
Sheng Jin, Lumin Xu, Jin Xu, Can Wang, Wentao Liu, Chen Qian, Wanli Ouyang, and
  Ping Luo.
\newblock Whole-body human pose estimation in the wild.
\newblock In {\em ECCV}, 2020.

\bibitem{inproceedings(LSP)}
Sam Johnson and Mark Everingham.
\newblock Clustered pose and nonlinear appearance models for human pose
  estimation.
\newblock In {\em BMVC}, 2010.

\bibitem{Joo_2015_ICCV(Panoptic)}
Hanbyul Joo, Hao Liu, Lei Tan, Lin Gui, Bart Nabbe, Iain Matthews, Takeo
  Kanade, Shohei Nobuhara, and Yaser Sheikh.
\newblock Panoptic studio: A massively multiview system for social motion
  capture.
\newblock In {\em ICCV}, 2015.

\bibitem{DBLP:journals/corr/abs-1801-01615(totalcapture)}
Hanbyul Joo, Tomas Simon, and Yaser Sheikh.
\newblock Total capture: {A} 3d deformation model for tracking faces, hands,
  and bodies.
\newblock {\em CVPR}, 2018.

\bibitem{kanazawa2018end}
Angjoo Kanazawa, Michael~J Black, David~W Jacobs, and Jitendra Malik.
\newblock End-to-end recovery of human shape and pose.
\newblock In {\em CVPR}, 2018.

\bibitem{kolotouros2019learning}
Nikos Kolotouros, Georgios Pavlakos, Michael~J. Black, and Kostas Daniilidis.
\newblock Learning to reconstruct 3d human pose and shape via model-fitting in
  the loop.
\newblock {\em ICCV}, 2019.

\bibitem{kolotouros2019convolutional}
Nikos Kolotouros, Georgios Pavlakos, and Kostas Daniilidis.
\newblock Convolutional mesh regression for single-image human shape
  reconstruction.
\newblock In {\em CVPR}, 2019.

\bibitem{kreiss2021openpifpaf}
Sven Kreiss, Lorenzo Bertoni, and Alexandre Alahi.
\newblock {OpenPifPaf: Composite Fields for Semantic Keypoint Detection and
  Spatio-Temporal Association}.
\newblock {\em IEEE Transactions on Intelligent Transportation Systems}, 2021.

\bibitem{kulon2020weakly}
Dominik Kulon, Riza~Alp Guler, Iasonas Kokkinos, Michael~M Bronstein, and
  Stefanos Zafeiriou.
\newblock Weakly-supervised mesh-convolutional hand reconstruction in the wild.
\newblock In {\em CVPR}, 2020.

\bibitem{lassner2017unite}
Christoph Lassner, Javier Romero, Martin Kiefel, Federica Bogo, Michael~J
  Black, and Peter~V Gehler.
\newblock Unite the people: Closing the loop between 3d and 2d human
  representations.
\newblock In {\em CVPR}, 2017.

\bibitem{Liu_2020_CVPR(PoseReconstruction)}
Ruixu Liu, Ju Shen, He Wang, Chen Chen, Sen-ching Cheung, and Vijayan Asari.
\newblock Attention mechanism exploits temporal contexts: Real-time 3d human
  pose reconstruction.
\newblock In {\em CVPR}, 2020.

\bibitem{SMPL:2015(SMPLModel)}
Matthew Loper, Naureen Mahmood, Javier Romero, Gerard Pons-Moll, and Michael~J.
  Black.
\newblock {SMPL}: A skinned multi-person linear model.
\newblock {\em ACM Trans. Graphics (Proc. SIGGRAPH Asia)}, 2015.

\bibitem{https://doi.org/10.48550/arxiv.2208.03704(jointformer)}
Sebastian Lutz, Richard Blythman, Koustav Ghosal, Matthew Moynihan, Ciaran
  Simms, and Aljosa Smolic.
\newblock Jointformer: Single-frame lifting transformer with error prediction
  and refinement for 3d human pose estimation.
\newblock {\em ArXiv}, 2022.

\bibitem{DBLP:journals/corr/MartinezHRL17}
Julieta Martinez, Rayat Hossain, Javier Romero, and James~J. Little.
\newblock A simple yet effective baseline for 3d human pose estimation.
\newblock In {\em ICCV}, 2017.

\bibitem{mehta2017monocular(3DHP)}
Dushyant Mehta, Helge Rhodin, Dan Casas, Pascal Fua, Oleksandr Sotnychenko,
  Weipeng Xu, and Christian Theobalt.
\newblock Monocular 3d human pose estimation in the wild using improved cnn
  supervision.
\newblock In {\em 3DV}, 2017.

\bibitem{mehta2018single}
Dushyant Mehta, Oleksandr Sotnychenko, Franziska Mueller, Weipeng Xu, Srinath
  Sridhar, Gerard Pons-Moll, and Christian Theobalt.
\newblock Single-shot multi-person 3d pose estimation from monocular rgb.
\newblock In {\em 3DV}, 2018.

\bibitem{moon2019camera}
Gyeongsik Moon, Ju~Yong Chang, and Kyoung~Mu Lee.
\newblock Camera distance-aware top-down approach for 3d multi-person pose
  estimation from a single {RGB} image.
\newblock {\em ICCV}, 2019.

\bibitem{DBLP:journals/corr/abs-2008-09309(hand3)}
Gyeongsik Moon, Shoou{-}I Yu, He Wen, Takaaki Shiratori, and Kyoung~Mu Lee.
\newblock Interhand2.6m: {A} dataset and baseline for 3d interacting hand pose
  estimation from a single {RGB} image.
\newblock {\em ECCV}, 2020.

\bibitem{mueller2018ganerated}
Franziska Mueller, Florian Bernard, Oleksandr Sotnychenko, Dushyant Mehta,
  Srinath Sridhar, Dan Casas, and Christian Theobalt.
\newblock Ganerated hands for real-time 3d hand tracking from monocular rgb.
\newblock In {\em CVPR}, 2018.

\bibitem{DBLP:journals/corr/NewellYD16_(stackhourglass)}
Alejandro Newell, Kaiyu Yang, and Jia Deng.
\newblock Stacked hourglass networks for human pose estimation.
\newblock {\em ECCV}, 2016.

\bibitem{smplifyx}
Georgios Pavlakos, Vasileios Choutas, Nima Ghorbani, Timo Bolkart, Ahmed Osman,
  Dimitrios Tzionas, and Michael Black.
\newblock Expressive body capture: {3D} hands, face, and body from a single
  image.
\newblock {\em CVPR}, 2019.

\bibitem{DBLP:journals/corr/PavlakosZDD16(CoarsetoFine)}
Georgios Pavlakos, Xiaowei Zhou, Konstantinos~G. Derpanis, and Kostas
  Daniilidis.
\newblock Coarse-to-fine volumetric prediction for single-image 3d human pose.
\newblock {\em CVPR}, 2017.

\bibitem{petrovich22temos}
Mathis Petrovich, Michael~J. Black, and G{\"u}l Varol.
\newblock {TEMOS}: Generating diverse human motions from textual descriptions.
\newblock In {\em European Conference on Computer Vision ({ECCV})}, 2022.

\bibitem{lcrnet}
Gr{\'{e}}gory Rogez, Philippe Weinzaepfel, and Cordelia Schmid.
\newblock Lcr-net++: Multi-person 2d and 3d pose detection in natural images.
\newblock {\em TPAMI}, 2019.

\bibitem{romero2022embodied}
Javier Romero, Dimitrios Tzionas, and Michael~J Black.
\newblock Embodied hands: Modeling and capturing hands and bodies together.
\newblock {\em arXiv preprint arXiv:2201.02610}, 2022.

\bibitem{rong2022chasing}
Yu Rong, Ziwei Liu, and Chen~Change Loy.
\newblock Chasing the tail in monocular 3d human reconstruction with prototype
  memory.
\newblock {\em TIP}, 2022.

\bibitem{franmocap}
Yu Rong, Takaaki Shiratori, and Hanbyul Joo.
\newblock Frankmocap: {A} monocular 3d whole-body pose estimation system via
  regression and integration.
\newblock {\em ICCV}, 2021.

\bibitem{saitoCVPR2021}
Shunsuke Saito, Jinlong Yang, Qianli Ma, and Michael~J. Black.
\newblock {SCANimate}: Weakly supervised learning of skinned clothed avatar
  networks.
\newblock In {\em Proceedings IEEE/CVF Conf.~on Computer Vision and Pattern
  Recognition (CVPR)}, June 2021.

\bibitem{sanyal2019learning}
Soubhik Sanyal, Timo Bolkart, Haiwen Feng, and Michael~J Black.
\newblock Learning to regress 3d face shape and expression from an image
  without 3d supervision.
\newblock In {\em CVPR}, 2019.

\bibitem{tuan2017regressing}
Anh Tuan~Tran, Tal Hassner, Iacopo Masi, and G{\'e}rard Medioni.
\newblock Regressing robust and discriminative 3d morphable models with a very
  deep neural network.
\newblock In {\em CVPR}, 2017.

\bibitem{DBLP:journals/corr/VaswaniSPUJGKP17(transformer)}
Ashish Vaswani, Noam Shazeer, Niki Parmar, Jakob Uszkoreit, Llion Jones,
  Aidan~N. Gomez, Lukasz Kaiser, and Illia Polosukhin.
\newblock Attention is all you need.
\newblock {\em NIPS}, 2017.

\bibitem{Marcard_2018_ECCV(3DPW)}
Timo von Marcard, Roberto Henschel, Michael~J. Black, Bodo Rosenhahn, and
  Gerard Pons-Moll.
\newblock Recovering accurate 3d human pose in the wild using imus and a moving
  camera.
\newblock In {\em ECCV}, 2018.

\bibitem{DBLP:journals/corr/abs-2011-14679(canonpose)}
Bastian Wandt, Marco Rudolph, Petrissa Zell, Helge Rhodin, and Bodo Rosenhahn.
\newblock Canonpose: Self-supervised monocular 3d human pose estimation in the
  wild.
\newblock In {\em CVPR}, 2021.

\bibitem{wang2021scene}
Jingbo Wang, Sijie Yan, Bo Dai, and Dahua Lin.
\newblock Scene-aware generative network for human motion synthesis.
\newblock In {\em CVPR}, 2021.

\bibitem{wang2020motion}
Jingbo Wang, Sijie Yan, Yuanjun Xiong, and Dahua Lin.
\newblock Motion guided 3d pose estimation from videos.
\newblock In {\em ECCV}, 2020.

\bibitem{dope}
Philippe Weinzaepfel, Romain Br{\'{e}}gier, Hadrien Combaluzier, Vincent Leroy,
  and Gr{\'{e}}gory Rogez.
\newblock {DOPE:} distillation of part experts for whole-body 3d pose
  estimation in the wild.
\newblock {\em ECCV}, 2020.

\bibitem{DBLP:journals/corr/abs-2110-04800(face4)}
Yandong Wen, Weiyang Liu, Bhiksha Raj, and Rita Singh.
\newblock Self-supervised 3d face reconstruction via conditional estimation.
\newblock {\em ICCV}, 2021.

\bibitem{xiang2019monocular}
Donglai Xiang, Hanbyul Joo, and Yaser Sheikh.
\newblock Monocular total capture: Posing face, body, and hands in the wild.
\newblock In {\em CVPR}, 2019.

\bibitem{Xu_2020_CVPR(Kinematic)}
Jingwei Xu, Zhenbo Yu, Bingbing Ni, Jiancheng Yang, Xiaokang Yang, and Wenjun
  Zhang.
\newblock Deep kinematics analysis for monocular 3d human pose estimation.
\newblock In {\em CVPR}, 2020.

\bibitem{yang2019aligning}
Linlin Yang, Shile Li, Dongheui Lee, and Angela Yao.
\newblock Aligning latent spaces for 3d hand pose estimation.
\newblock In {\em ICCV}, 2019.

\bibitem{Zanfir20eccv}
Andrei Zanfir, Eduard~Gabriel Bazavan, Hongyi Xu, Bill Freeman, Rahul
  Sukthankar, and Cristian Sminchisescu.
\newblock Weakly supervised 3d human pose and shape reconstruction with
  normalizing flows.
\newblock In {\em ECCV}, 2020.

\bibitem{Zanfir19aaai}
Mihai Zanfir, Elisabeta Oneata, Alin{-}Ionut Popa, Andrei Zanfir, and Cristian
  Sminchisescu.
\newblock Human synthesis and scene compositing.
\newblock {\em AAAI}, 2019.

\bibitem{zhang2019end}
Xiong Zhang, Qiang Li, Hong Mo, Wenbo Zhang, and Wen Zheng.
\newblock End-to-end hand mesh recovery from a monocular rgb image.
\newblock In {\em ICCV}, pages 2354--2364, 2019.

\bibitem{zhou2021monocular}
Yuxiao Zhou, Marc Habermann, Ikhsanul Habibie, Ayush Tewari, Christian
  Theobalt, and Feng Xu.
\newblock Monocular real-time full body capture with inter-part correlations.
\newblock In {\em CVPR}, 2021.

\bibitem{zimmermann2017learning}
Christian Zimmermann and Thomas Brox.
\newblock Learning to estimate 3d hand pose from single {RGB} images.
\newblock {\em ICCV}, 2017.

\bibitem{DBLP:journals/corr/abs-1909-04349(hand5)}
Christian Zimmermann, Duygu Ceylan, Jimei Yang, Bryan~C. Russell, Max Argus,
  and Thomas Brox.
\newblock Freihand: {A} dataset for markerless capture of hand pose and shape
  from single {RGB} images.
\newblock {\em ICCV}, 2019.

\end{thebibliography}
}
\newpage
\clearpage
\appendix

\section{Supplementary material -- Overview}
In this supplementary material:
\begin{itemize}
\item We share the link to download the H3WB dataset annotation files~(Section~\ref{sec:codeannotation}); 
\item We provide the H3WB 3D whole-body dataset keypoint layout with 133 keypoints. H3WB dataset follows exactly the same layout as COCO WholeBody dataset~\cite{jin2020whole(COCO-WholeBody)}~(Section~\ref{sec:layout}); 
\item We provide the statics regarding the diversity of H3WB dataset (Section~\ref{sec:diversity});
\item We present web interface of the quality assessment (Section~\ref{sec:crosscheck});
\item We provide more qualitative results for all tasks, as well as qualitative results in the wild evaluated on the COCO dataset (Section~\ref{sec:morevisuals});

\item We study failure cases from SMPL-X extracted from the literature (Section~\ref{sec:no_smplx});

\item We report the results of our
5-fold cross-validation experiments~(Section~\ref{sec:crossval});

\item We clarify long-term support planning and the license issue (Section~\ref{sec:non_technique_issues}).
\end{itemize}

\section{H3WB annotations}
\label{sec:codeannotation}
\noindent To download the H3WB dataset annotations click \href{https://drive.google.com/file/d/1O4qXYIcRuvcLXr_bMqIetpWpwTciDPER/view?usp=sharing}{here}. 
The zip file contains following:
\begin{itemize}
    \item 2Dto3D\_train.json has the training annotations for 2D$\rightarrow$3D and I2D$\rightarrow$3D tasks. Since this file is too big, we split it into 4-parts to ease the training and data loading pipeline. We provide the splitted files as well.
    \item RGBto3D\_train.json has the training annotations for RGB$\rightarrow$3D task.
    \item 2Dto3D\_test\_2d.json and I2Dto3D\_test\_2d.json include test instances for 2D$\rightarrow$3D and I2D$\rightarrow$3D tasks, respectively.
    \item RGBto3D\_test\_img.json includes test samples for RGB$\rightarrow$3D task.

\end{itemize}

\section{H3WB dataset keypoint layout}
\label{sec:layout}

\begin{figure}[h]
  \centering
  \includegraphics[width=0.8\linewidth]{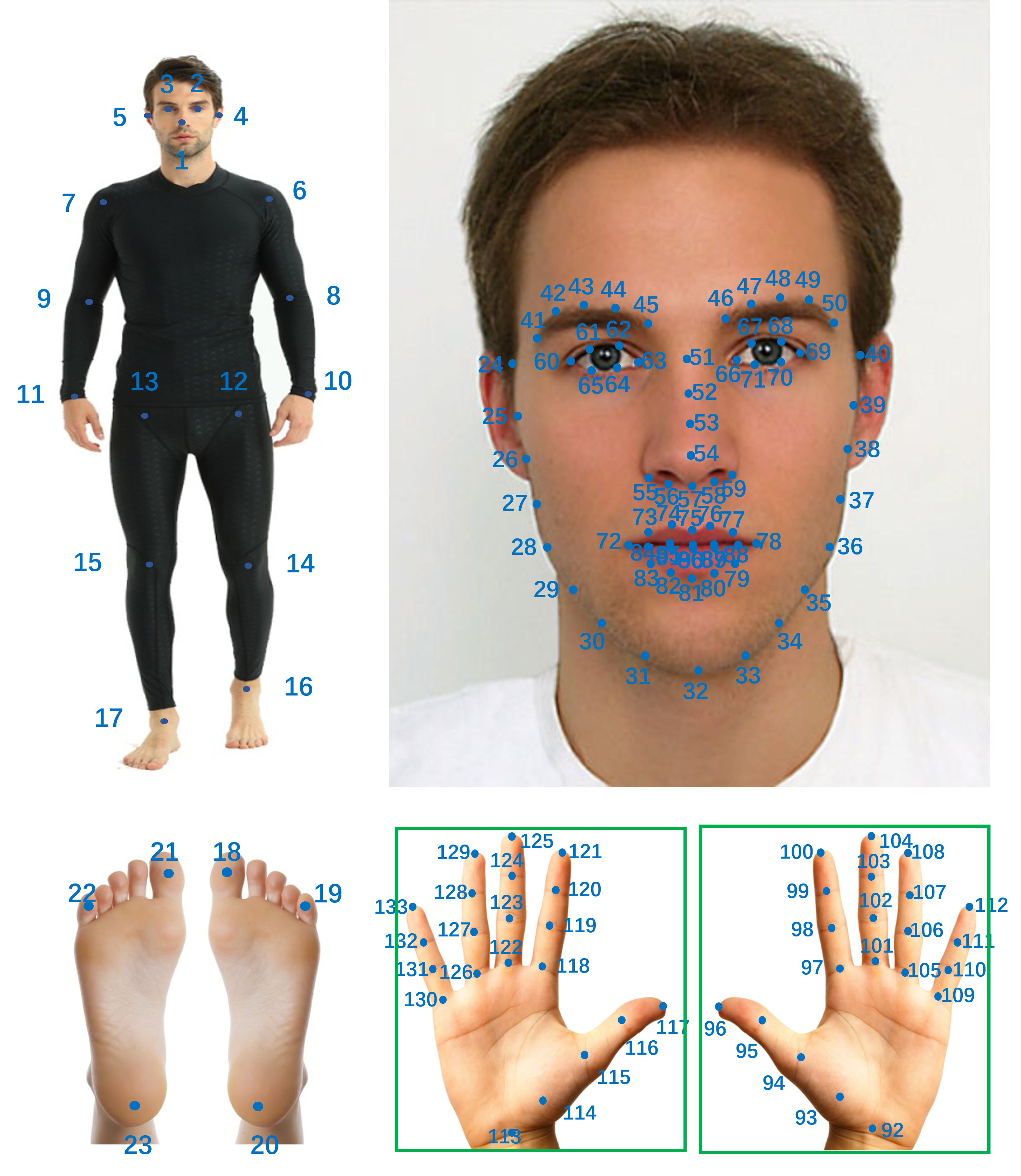}
  \caption{Whole-body keypoint layout defined in the COCO WholeBody dataset~\cite{jin2020whole(COCO-WholeBody)}. H3WB dataset follows exactly the same layout. H3WB dataset has total of 133 keypoints annotations for each human: 17 human body keypoints (top-left), 68 face (top-right), 42 hand (21 keypoints for each) (bottom-right) and 6 foot (3 for each) (bottom-left). Image source: https://github.com/jin-s13/COCO-WholeBody}
  \label{fig:layout}
\end{figure}

We use the COCO WholeBody dataset layout with 133 keypoints illustrated in Figure~\ref{fig:layout}. H3WB dataset has the same keypoints for the whole-body layout. 

\begin{figure}

  \centering
  \includegraphics[width=0.9\linewidth]{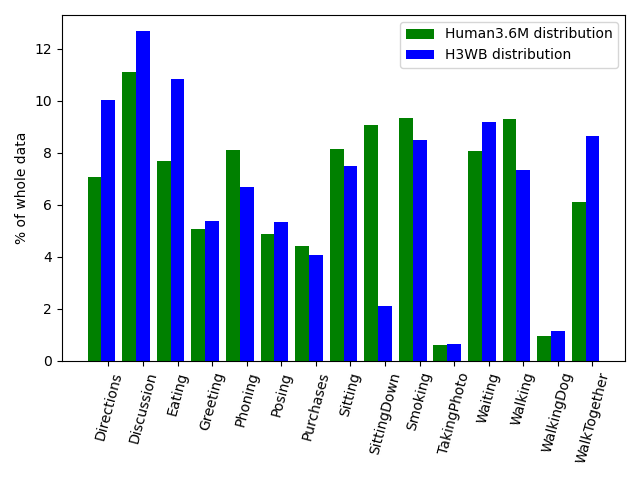}
  \caption{Distributions of Human3.6 and H3WB datasets per action class}
  \label{fig:diversity}

\end{figure}

\begin{table*}[h]
    \centering
    \resizebox{\linewidth}{!}{
    \begin{tabular}{l | l  lllllllllllllllll}
    \toprule
    H36M & \textbf{602.7} & 540.0 & 576.3 & 569.3 & 578.7 & 512.8 & 513.3 & 527.7 & 545.3 & 551.3 & 552.8 & 556.5 & 525.7 & 518.9 & 534.8 &    584.5 & 624.1 & 637.4 \\
    H3WB & \textbf{518.8} & 437.9 & 433.5 & 444.3 & 428.9 & 453.6 & 422.3 & 473.1 & 427.5 & 505.0 & 440.2 & 519.1 & 419.9 & 456.1 & 430.5 &    462.6 & 440.3 & 473.1\\
    \bottomrule
    \end{tabular}
    }
\caption{Standard deviation in mm on average (1st column) and for each of the original 17 body joints.}
    \label{tab:variance}
\end{table*}

\begin{figure*}
  \centering
\begin{tabular}{cc}
  \includegraphics[width=0.35\linewidth,valign=c]{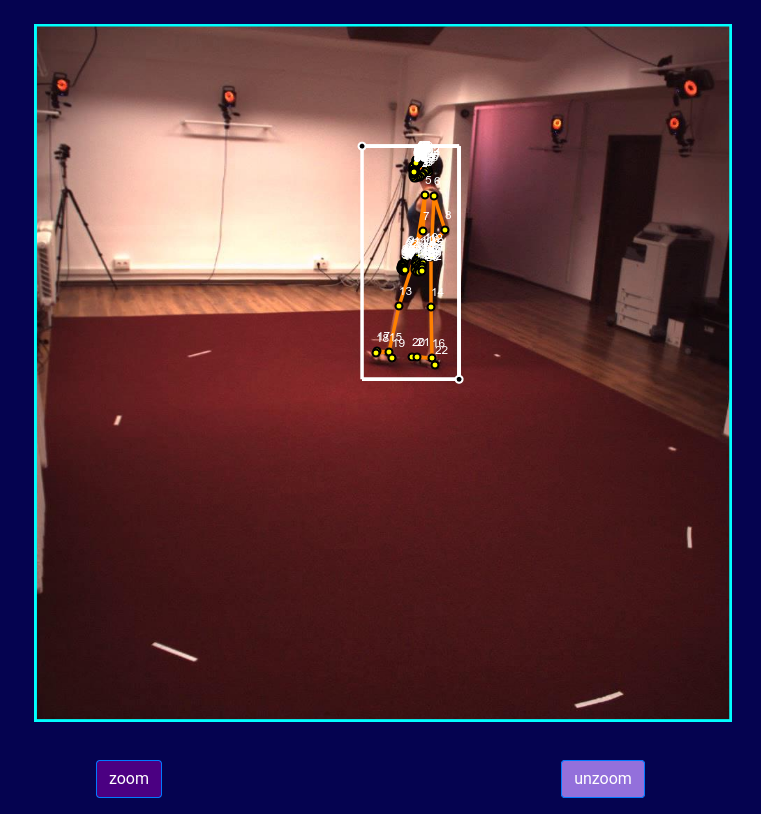} 
\includegraphics[width=0.35\linewidth,valign=c]{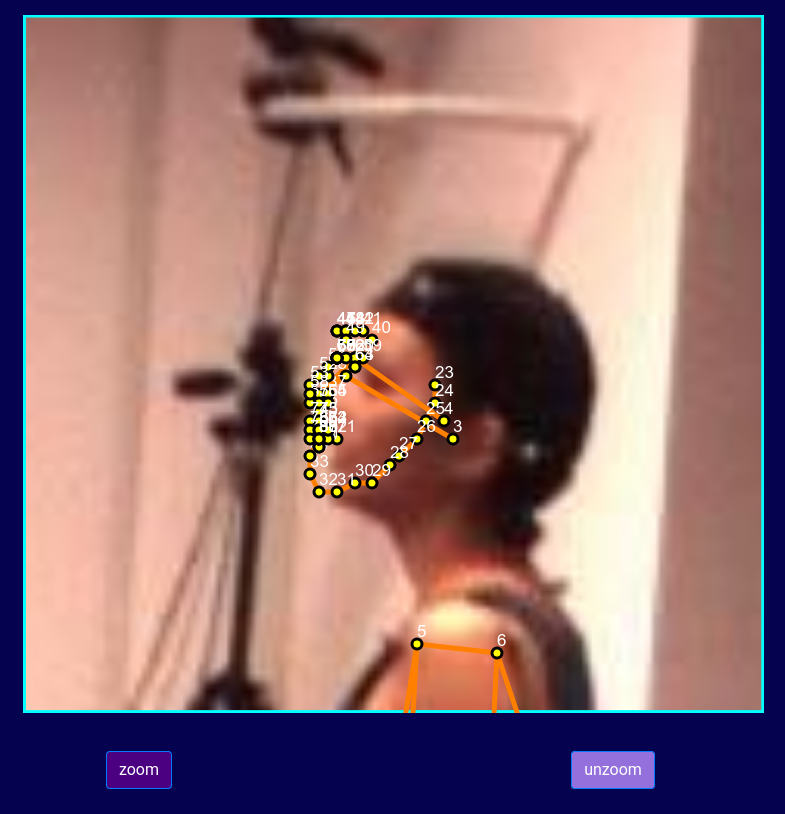} \\
\includegraphics[width=0.35\linewidth,valign=c]{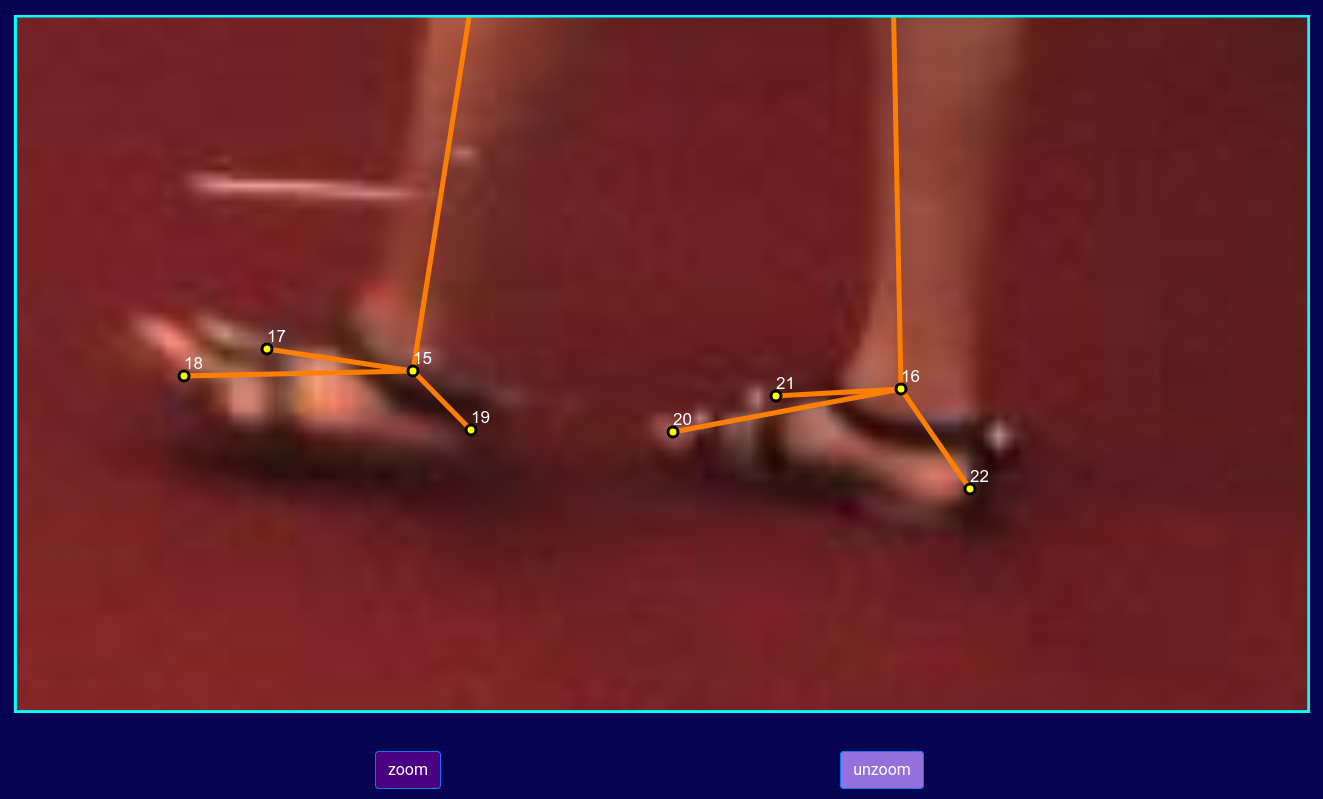} 
\includegraphics[width=0.35\linewidth, valign=c]{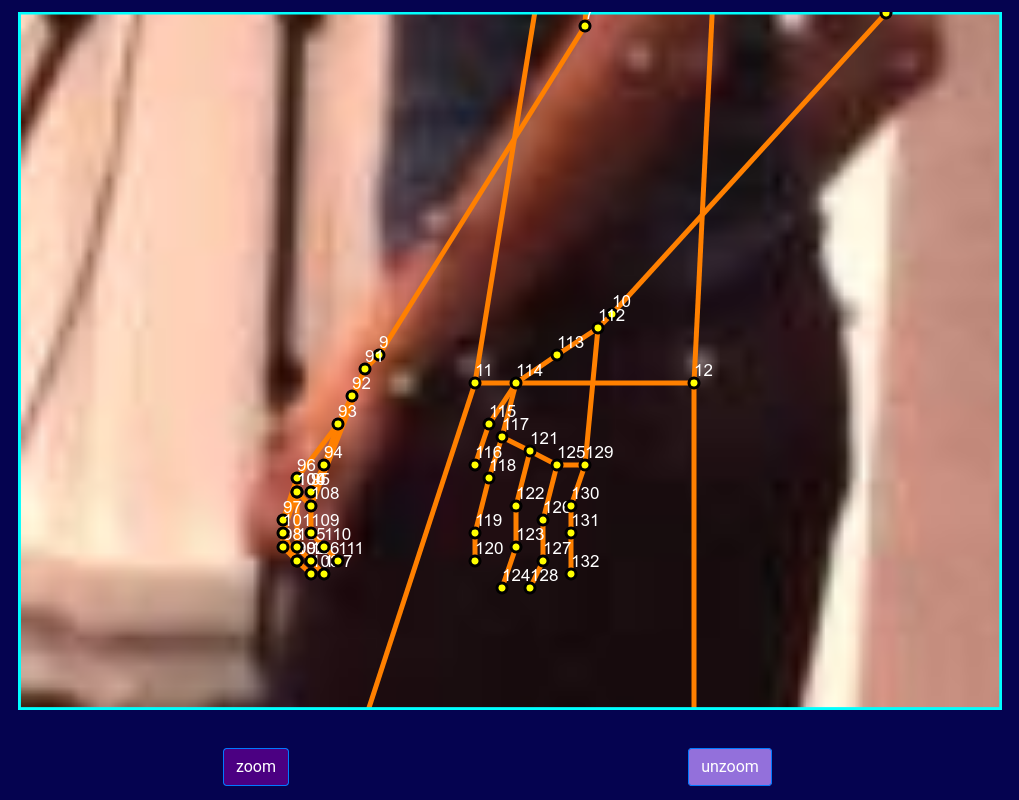}
\end{tabular}
  \caption{Sample screenshots from the annotation interface. Annotators are asked to select area of interest, zoom in on that area and correct the mis-aligned keypoints by drag-drop.}
  \label{fig:userstudy}
\end{figure*}

\section{Dataset diversity}
\label{sec:diversity}

The distribution of pose per action for H36M and H3WB using the original action labels is shown in \autoref{fig:diversity}. Apart from \emph{SittingDown}, they are about the same. Quantitatively, we show the standard deviation in mm on average (bold) and for each of the original 17 body joints in \autoref{tab:variance} which shows H3WB has slightly lower diversity than H36M, but no collapse.

\section{Quality assessment study}
\label{sec:crosscheck}

We assessed the quality of the H3WB dataset by manually annotating 80K keypoints from 600 randomly selected images from the dataset. We presented a web interface to annotators and ask them to zoom-in on the body parts and correct mis-aligned keypoints by drag and drop. Sample screenshots from our web interface are presented in Figure~\ref{fig:userstudy}.

\section{More Qualitative Results}
\label{sec:morevisuals}

\begin{figure*}
  \centering
  \includegraphics[width=0.26\linewidth]{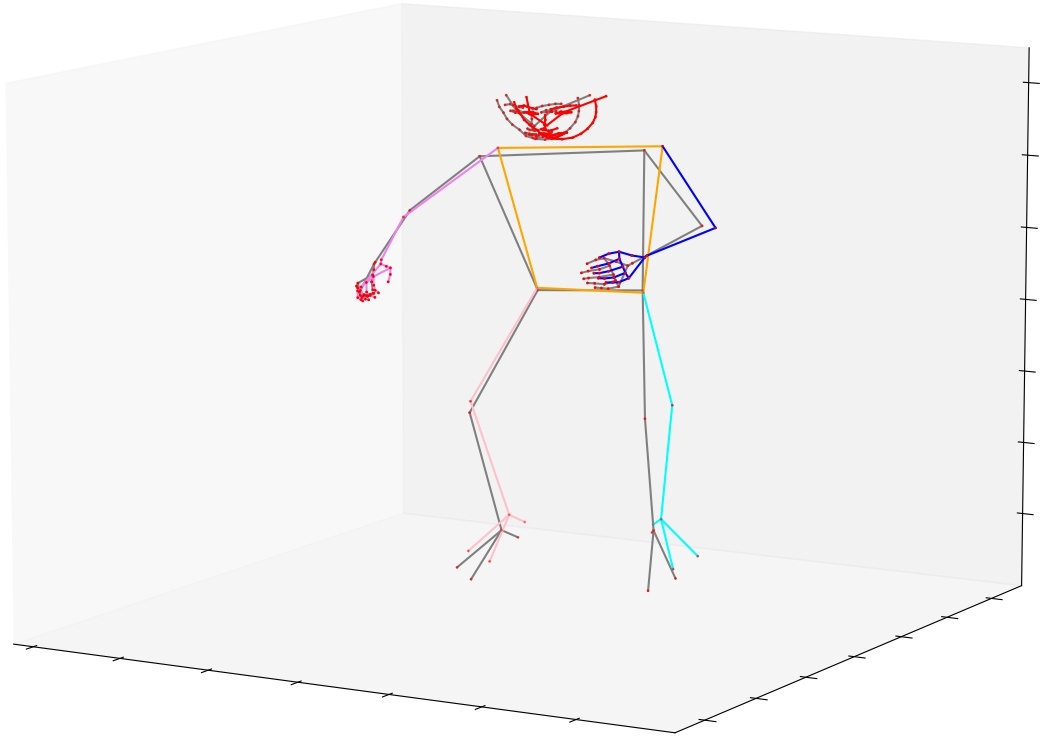} 
\includegraphics[width=0.26\linewidth]{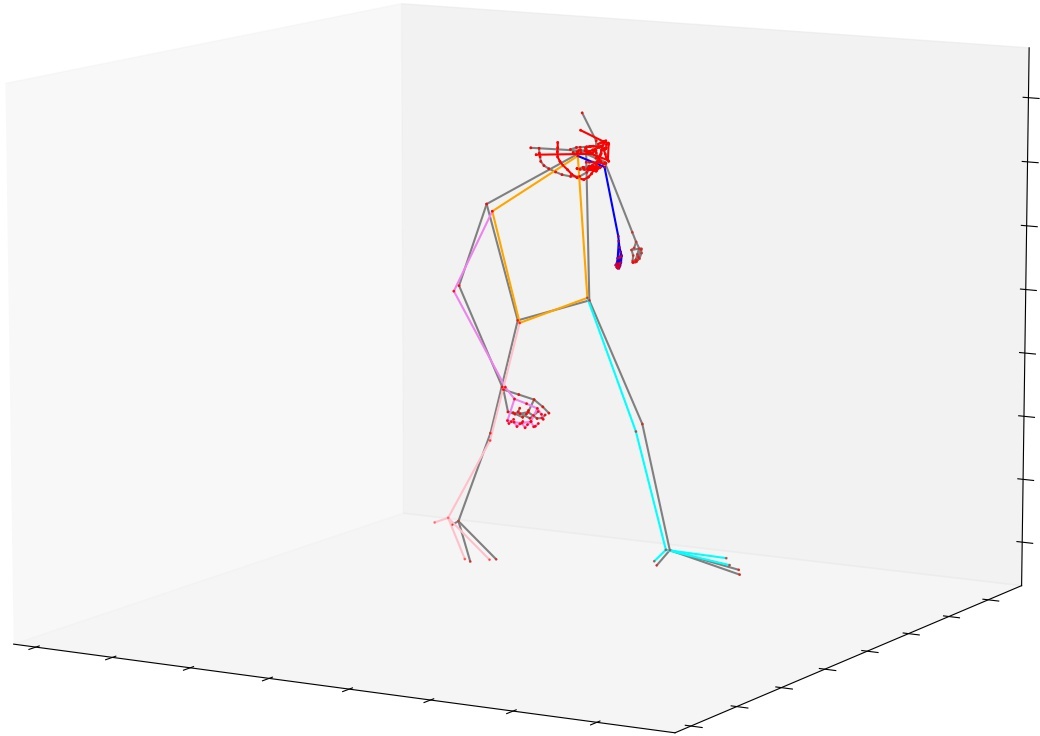} 
\includegraphics[width=0.26\linewidth]{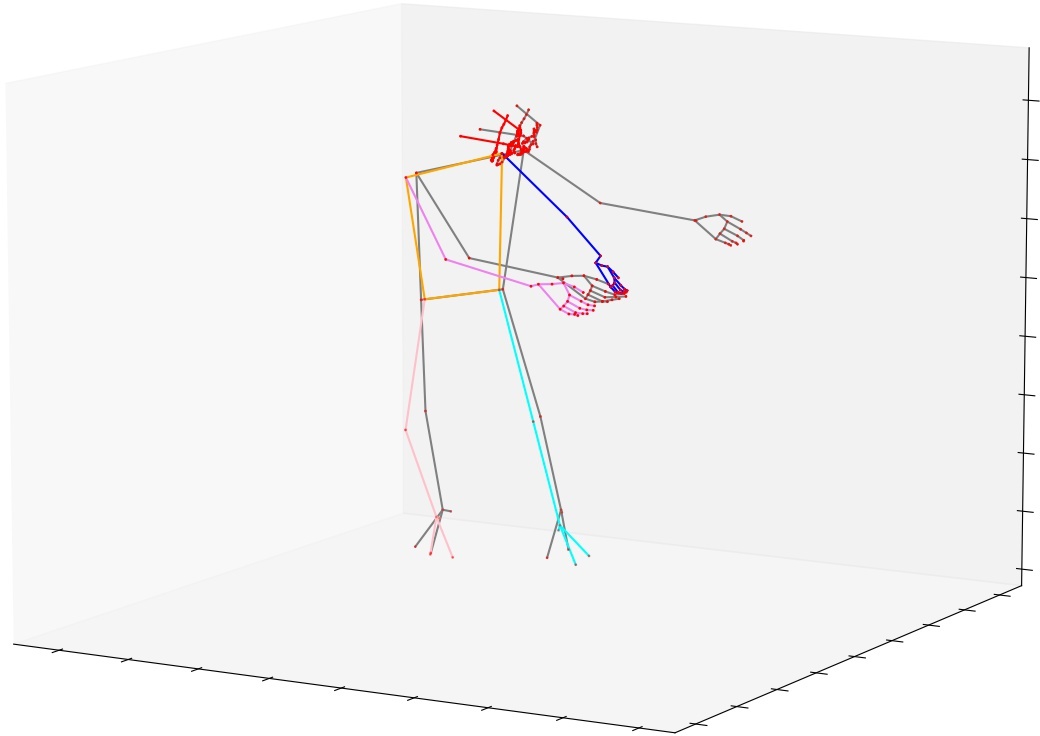}  \\
 \includegraphics[width=0.26\linewidth]{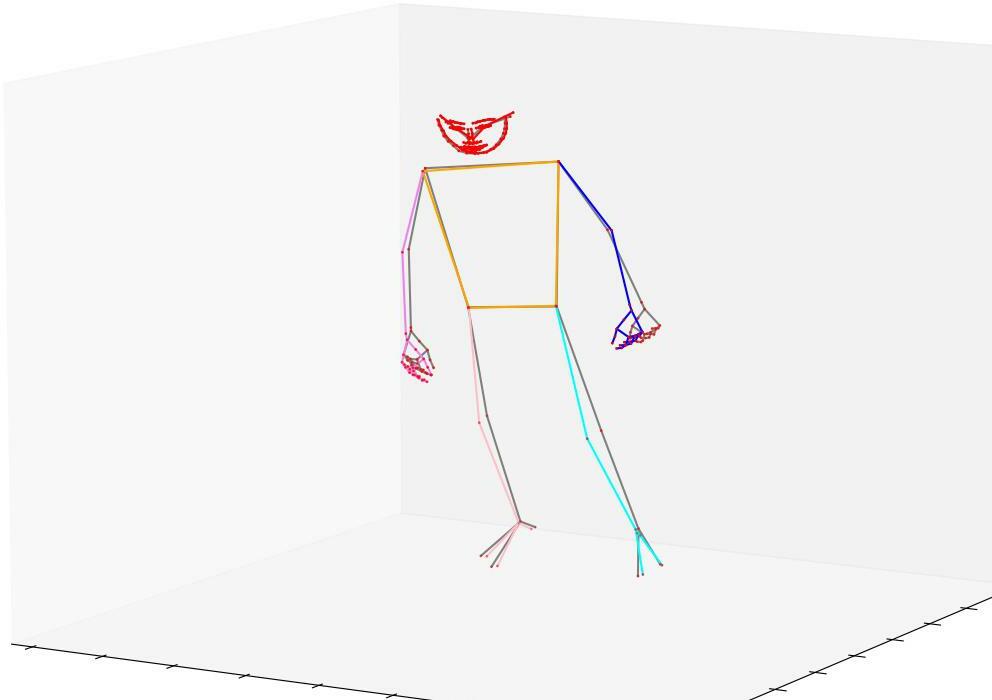} 
\includegraphics[width=0.26\linewidth]{img/t2m2/outputboth-20.jpg} 
\includegraphics[width=0.26\linewidth]{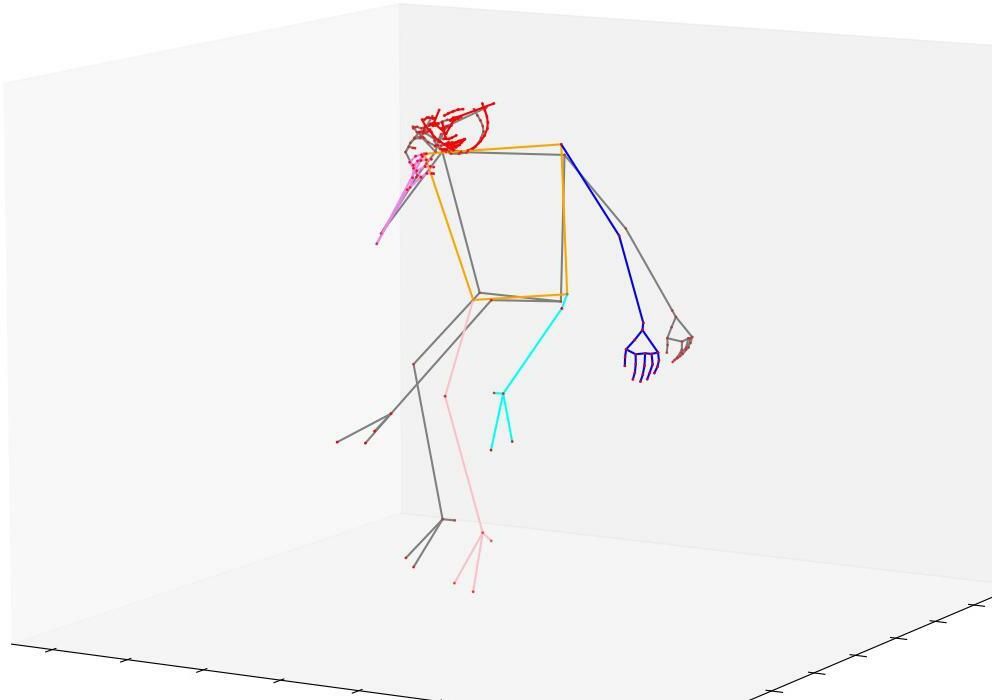}  \\
 \includegraphics[width=0.26\linewidth]{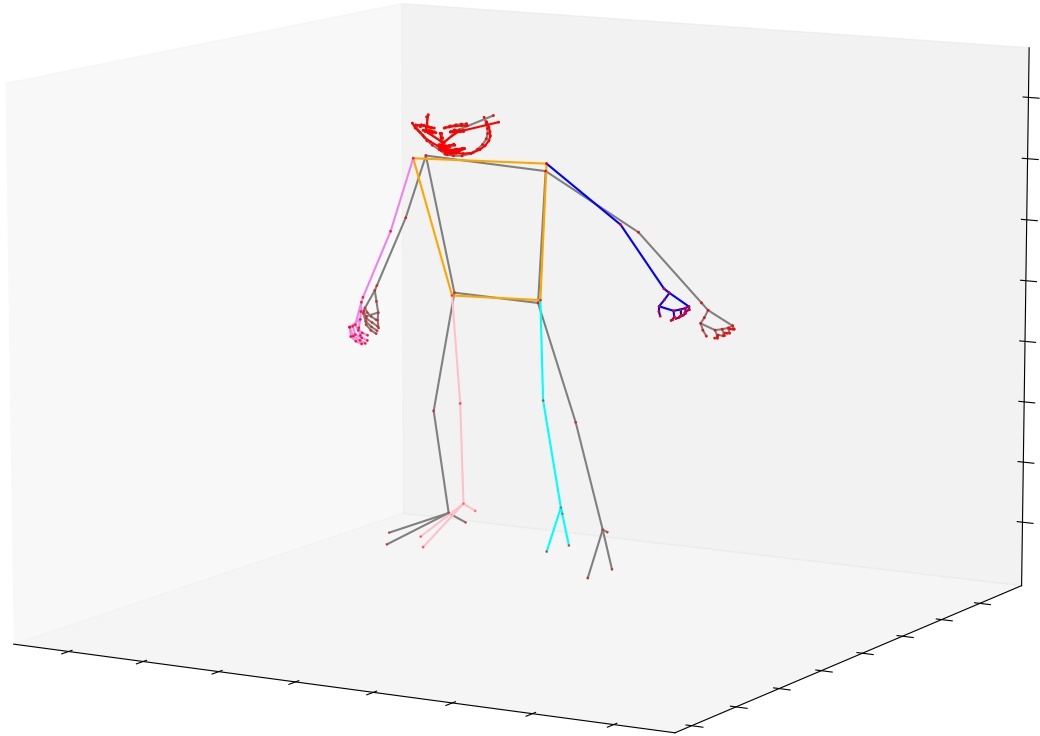} 
\includegraphics[width=0.26\linewidth]{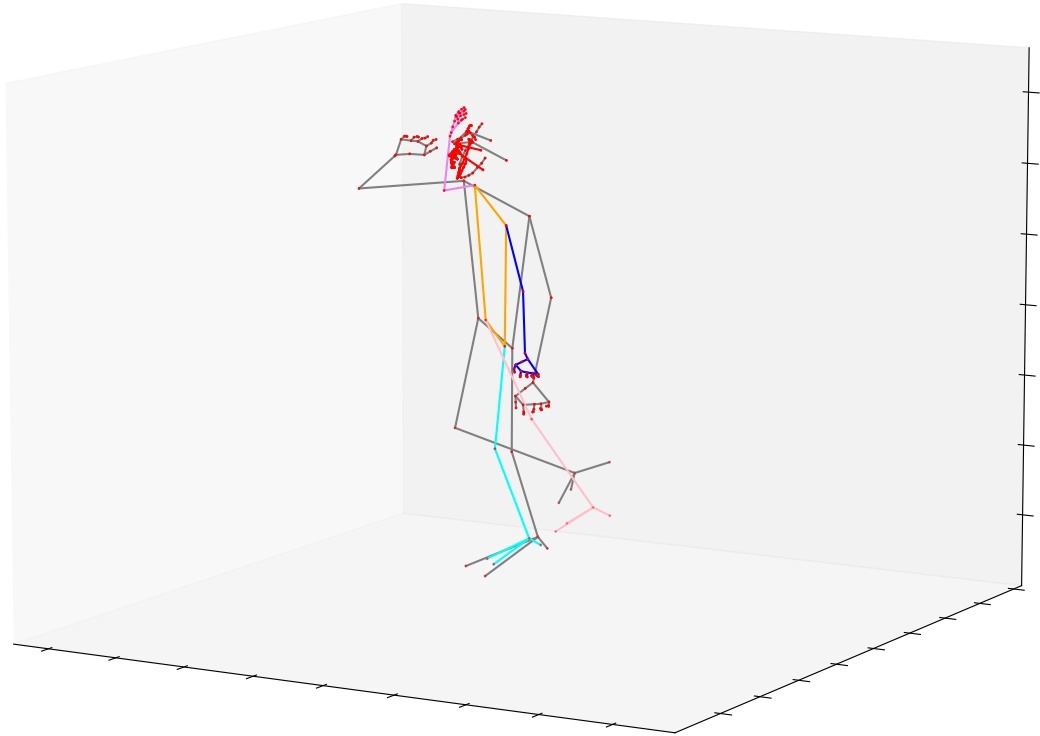} 
\includegraphics[width=0.26\linewidth]{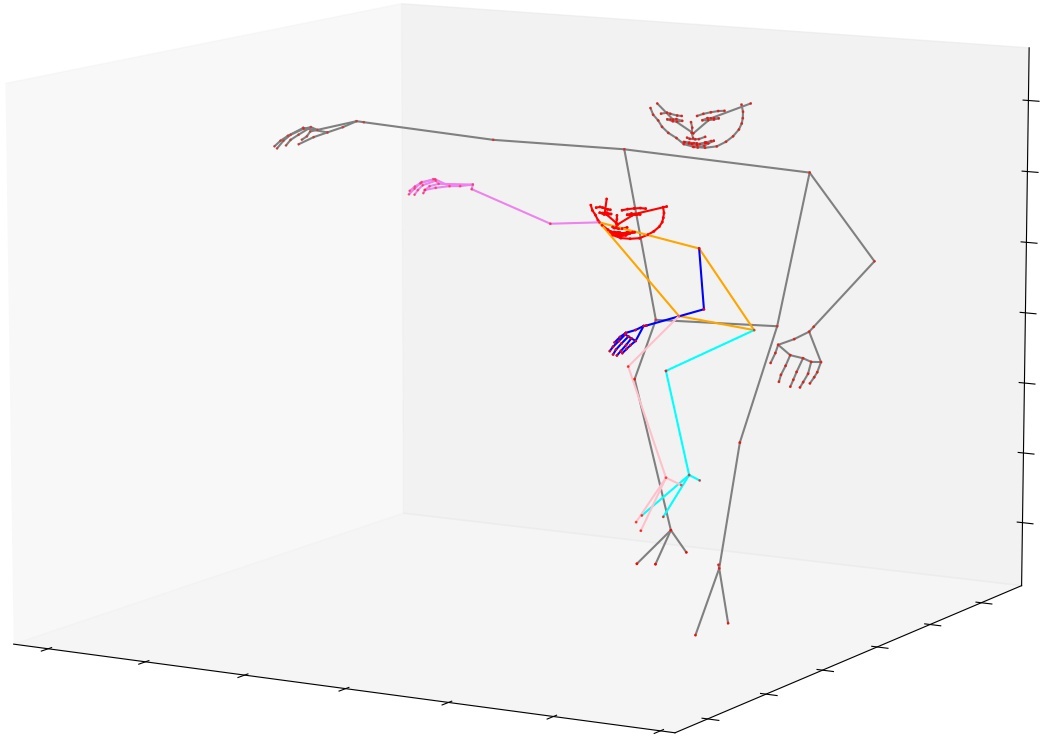}  \\

  \caption{Example predictions from Large SimpleBaseline model for 2D$\rightarrow$3D (1st row) and I2D$\rightarrow$3D (2nd row) tasks. 3rd row shows predictions from Jointformer for RGB$\rightarrow$3D task.  Colored skeletons correspond to predictions and gray skeletons correspond to  groundtruths. First two columns show almost-aligned successful front/side predictions, and the last column shows slightly mis-aligned predictions.}
  \label{fig:morevisual}
\end{figure*}

We provide more qualitative outputs obtained by Large SimpleBaseline~\cite{DBLP:journals/corr/MartinezHRL17} and Jointformer~\cite{https://doi.org/10.48550/arxiv.2208.03704(jointformer)} models in \autoref{fig:morevisual}. Despite slight mis-alignments, the predicted skeletons are realistic.

\begin{figure*}
  \centering
  \includegraphics[width=0.3\linewidth]{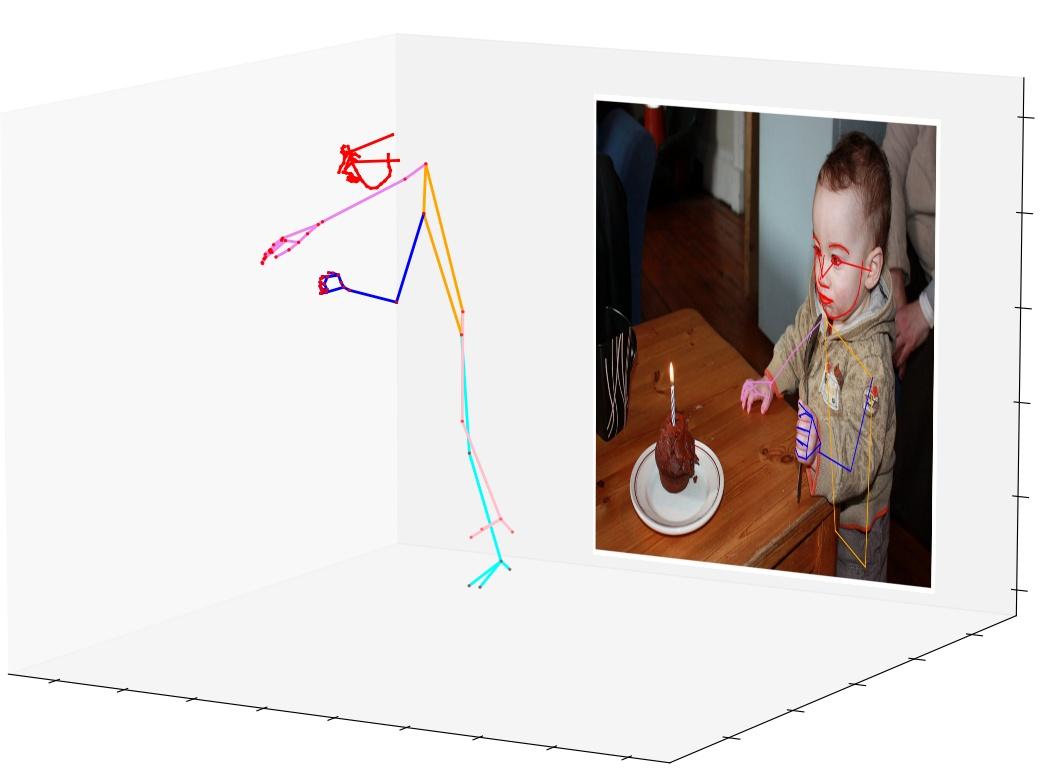}
  \includegraphics[width=0.3\linewidth]{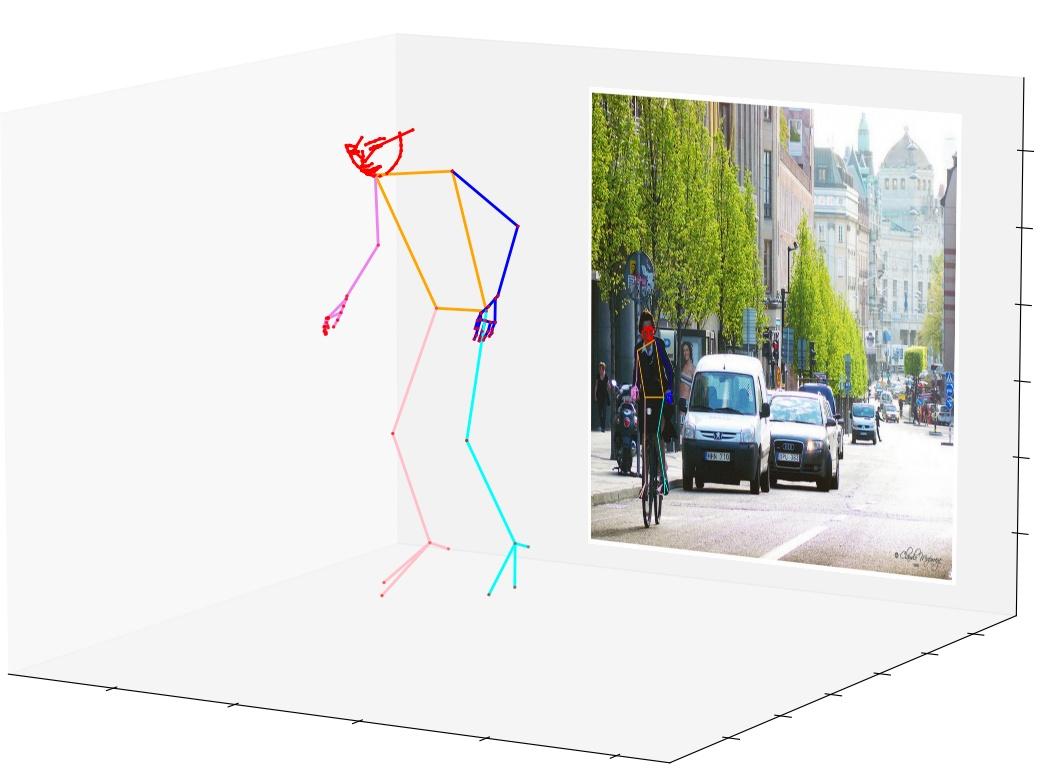}
  \includegraphics[width=0.3\linewidth]{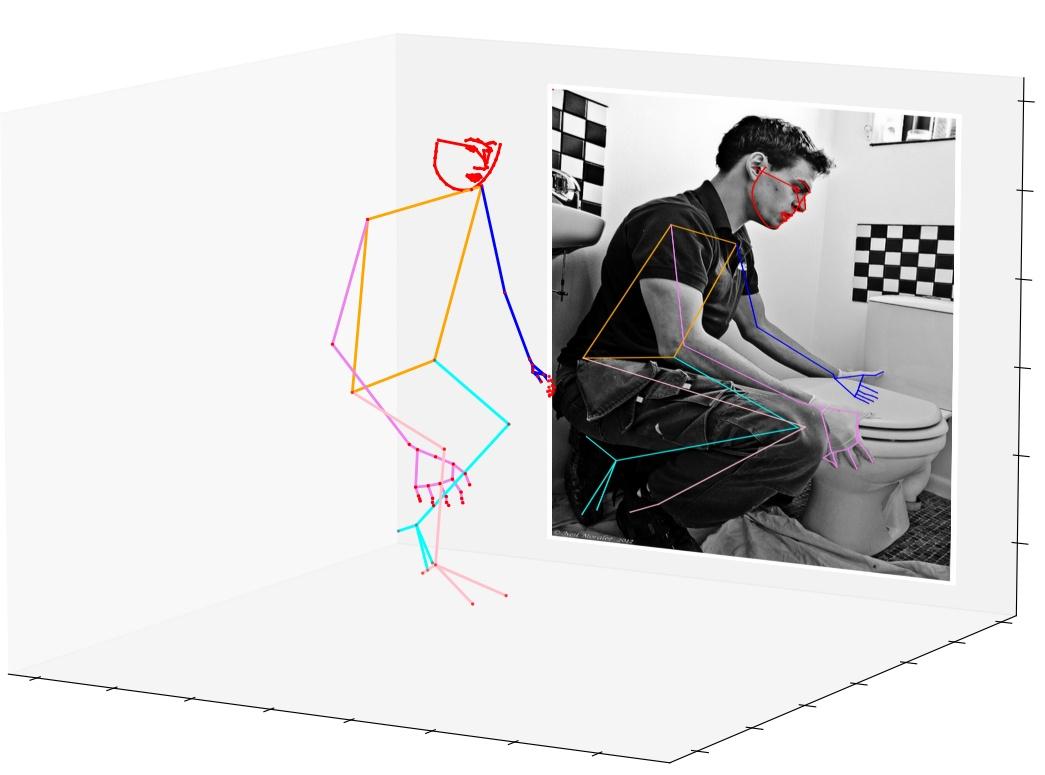}
  \caption{Visual examples of lifting on COCO. The labels on the images are the incomplete inputs.}
  \label{fig:coco-inference}
\end{figure*}

\begin{figure*}[t]
  \centering
  \includegraphics[width=\linewidth]{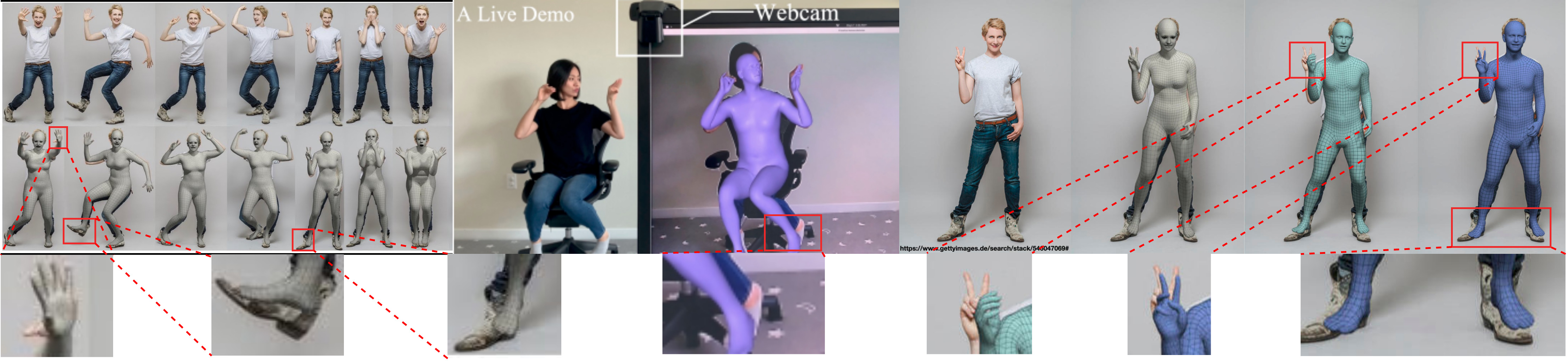} 
  \caption{Several examples of failures on hands and feet with SMPL-X model copied from \cite{smplifyx, franmocap, expose}}
  \label{fig:smplx_go_wrong}
\end{figure*}

We also show some examples in \autoref{fig:coco-inference} of a model trained on our H3WB benchmark for the task I2D$\rightarrow$3D and evaluated on COCO dataset\cite{jin2020whole(COCO-WholeBody)}. We can see that even when there are missing points in the 2D input, the model still can predict the 3D wholebody pose accurately. This validates the usefulness of the I2D$\rightarrow$3D in real world scenario.

\section{SMPL-X failure cases}
\label{sec:no_smplx}
\begin{figure}
  \centering
\includegraphics[width=\linewidth]{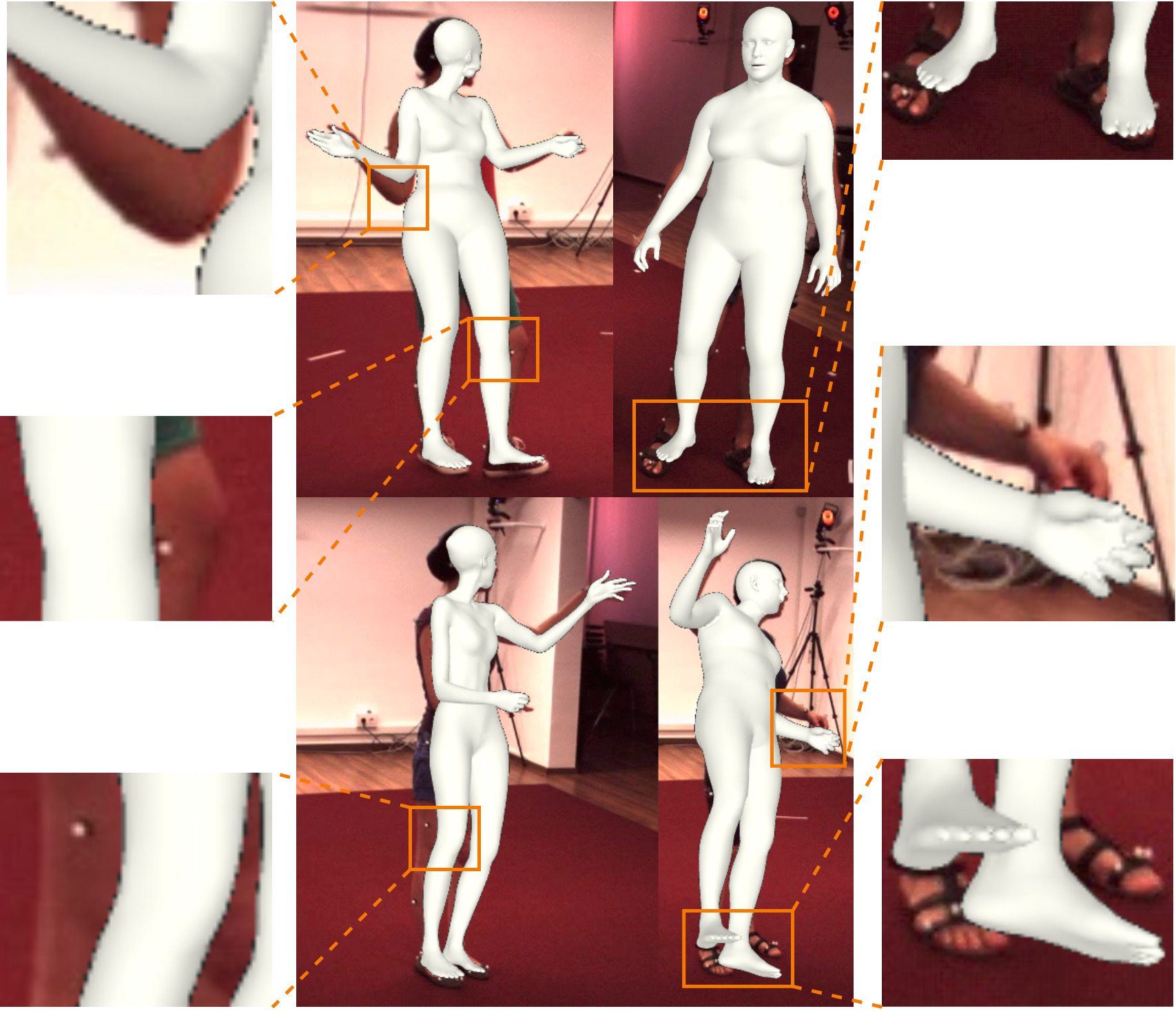}
  \caption{Our runs with SMPL-X models on the annotation, it is not as visually accurate as we require as annotations.}
  \label{fig:our_smplx}
\end{figure}

Parametric body models like SMPL-X have many seminal advantages such as always producing biologically plausible poses or taking into account the shape of the person. This enables powerful applications, for example in augmented reality or animation. However, because very accurate pose is not a requirement in these application, a model like SMPL-X is not yet able to reach satisfactory accuracy, especially on extremities like the hands and the feet. This is what we show in \autoref{fig:smplx_go_wrong}, where we extracted images from several articles \cite{smplifyx, franmocap, expose} and zoom on the extremities to visually assess that it is well below the accuracy provided in H3WB. We also ran SMPL-X on Human3.6M to see if it can be used to generate pseudo-labels and show selected zooms on the extremities on \autoref{fig:our_smplx}. Here also, the accuracy is well below what our label generation process managed to get. As such, datasets relying on SMPL-X for their groundtruth are thus by design less accurate and thus not usable for accurate pose estimation, especially on the extremities. Furthermore, assessing quantitatively the accuracy is almost impossible to do with these methods, whereas we provide an estimate for H3WB showing our benchmark is rigorous.

\section{Cross validation experiments}
\label{sec:crossval}

We do not provide a validation set for the H3WB dataset. We recommend 5-fold cross-validation for model selection and hyper-parameters tuning. 
We split the training set into 5  sets. We take the set cv$i$ as a hold out (test set), use remaining sets to train the models, and report the results on  cv$i$. We present the cross-validation results together with the test set results in Tables~\ref{tab:baseline_result1},~\ref{tab:baseline_result2},~\ref{tab:baseline_result3} for all tasks. We observe that cross-validation results are consistent and compatible with the test results which are listed in the main paper.

\begin{table}[t]
    \scriptsize
    \centering
    
    \begin{tabular}{l l l ll ll}
    \toprule
    \multicolumn{1}{c}{method} & \multicolumn{1}{c}{all} &
    \multicolumn{1}{c}{body} &
    \multicolumn{2}{c}{face} &
    \multicolumn{2}{c}{hand} \\
    \multicolumn{3}{c}{} &  & $\dagger$ &  & $\ddagger$ \\
    \midrule
    \textit{SimpleBaseline~\cite{DBLP:journals/corr/MartinezHRL17}} & & &  &   &   &   \\
    cv1 & 134.0 & 128.9 & 126.8 & 34.8 & 148.4 & 46.2 \\
    cv2 & 128.9 & 126.4 & 120.5 & 29.0 & 136.7 & 42.8 \\
    cv3 & 136.0 & 130.6 & 135.8 & 23.6 & 139.1 & 44.0 \\
    cv4 & 132.8 & 131.4 & 126.9 & 29.4 & 143.3 & 46.6 \\
    cv5 & 139.9 & 139.9 & 133.6 & 33.3 & 150.0 & 46.6 \\
    Cv std & 4.1 & 5.1 & 6.1 & 4.4 & 5.7 & 1.7 \\
    Cv mean & 134.3 & 131.4 & 128.7 &  30.0 & 143.5 &  45.2 \\
    Test & 125.4 & 125.7 & 115.9 & 24.6 & 140.7 & 42.5 \\

    \midrule
    \textit{Large SimpleBaseline~\cite{DBLP:journals/corr/MartinezHRL17}} & & & & & & \\
    cv1 & 106.8 & 105.1 & 105.8 & 22.7 & 109.2 & 33.6 \\
    cv2 & 103.9 & 104.3 & 107.7 & 16.7 & 97.6 & 31.9 \\
    cv3 & 101.8 & 102.4 & 105.5 & 14.2 & 95.6 & 30.6 \\
    cv4 & 108.7 & 107.0 & 111.6 & 14.3 & 105.0 & 32.3\\
    cv5 & 111.8 & 109.0 & 112.0 & 15.6 & 113.1 & 35.5 \\
    Cv std & 3.9 & 2.5 & 3.1 & 3.5 & 7.5 & 1.9 \\
    Cv mean & 106.6 & 105.6 & 108.5 &  16.7 & 104.1 & 32.8 \\
    Test & 112.3 & 112.6 & 110.6 & 14.6 & 114.8 & 31.7 \\
   \midrule
    \textit{CanonPose~\cite{DBLP:journals/corr/abs-2011-14679(canonpose)}} &   &   &  &  &   &   \\
    cv1 & 173.4 & 177.8 & 180.0 & 30.3 & 160.4 & 46.4 \\
    cv2 & 152.9 & 160.7 & 162.0 & 23.2 & 133.9 & 44.2 \\
    cv3 & 163.9 & 167.4 & 176.5 & 21.1 & 141.5 & 44.9 \\
    cv4 & 185.0 & 187.4 & 199.2 & 23.0 & 160.5 & 48.1 \\
    cv5 & 172.6 & 177.9 & 182.2 & 22.1 & 154.1 & 46.8 \\
    Cv std & 12.0 & 10.4 & 13.3 &  3.7 & 11.9 &  1.6 \\
    Cv mean & 169.6 & 174.2 & 180.0 &  23.9 & 150.1 &  46.1 \\
    Test & 186.7 & 193.7 & 188.4 & 24.6 & 180.2 & 48.9 \\
    \midrule
    \textit{CanonPose~\cite{DBLP:journals/corr/abs-2011-14679(canonpose)} + 3D sv.} &  & &  &  &  &  \\
    cv1 & 121.1 & 121.9 & 116.8 & 27.6 & 127.5 & 42.1 \\
    cv2 & 115.4 & 118.6 & 116.4 & 20.6 & 111.9 & 40.6 \\
    cv3 & 112.4 & 113.2 & 113.7 & 16.5 & 110.0 & 38.9 \\
    cv4 & 116.2 & 117.9 & 115.5 & 17.5 & 116.2 & 40.2 \\
    cv5 & 168.7 & 170.5 & 180.3 & 22.1 & 149.0 & 49.2 \\
    Cv std & 23.7 & 23.7 & 29.0 &  4.4 & 16.1 &  4.1 \\
    Cv mean & 126.8 & 128.4 & 128.5 &  20.9 & 122.9 &  42.2 \\
    Test & 117.7 & 117.5 & 112.0 & 17.9 & 126.9 & 38.3 \\
    \midrule
    \textit{Jointformer~\cite{https://doi.org/10.48550/arxiv.2208.03704(jointformer)}} & & & & & &  \\
    cv1 & 94.3 & 85.0 & 76.0 & 29.8 & 129.0 & 48.1 \\
    cv2 & 87.4 & 80.0 & 71.2 & 21.6 & 117.8 & 47.0 \\
    cv3 & 94.5 & 86.3 & 84.5 & 16.5 & 115.3 & 49.2 \\
    cv4 & 91.4 & 88.1 & 74.6 & 16.5 & 123.7 & 48.8 \\
    cv5 & 104.3 & 96.3 & 82.6 & 19.0 & 143.9 & 53.5 \\
    Cv std &  6.2 &  5.9 &  5.6 &  5.5 & 11.4 &  2.5 \\
    Cv mean &  94.4 &  87.1 &  77.8 &  20.7 & 125.9 &  49.3 \\
    Test & 88.3 & 84.9 & 66.5 & 17.8 & 125.3 & 43.7 \\

    \bottomrule
    \end{tabular}
    \caption{Results for 2D$\rightarrow$3D task on each 5-fold and test sets. Results are shown for MPJPE metric. All results are pelvis aligned, except $\dagger$ and $\ddagger$ show nose and wrist aligned results for face and hands, respectively. Sv. is supervision.}
    \label{tab:baseline_result1}
\end{table}

\begin{table}[t]
    \scriptsize
    \centering
    \begin{tabular}{l l l ll ll}
    \toprule
    \multicolumn{1}{c}{method} & \multicolumn{1}{c}{all} &
    \multicolumn{1}{c}{body} &
    \multicolumn{2}{c}{face} &
    \multicolumn{2}{c}{hand} \\
    \multicolumn{3}{c}{} &  & $\dagger$ &   & $\ddagger$ \\
    \midrule
    \textit{SimpleBaseline~\cite{DBLP:journals/corr/MartinezHRL17}} & &  &  &  &  & \\
    cv1 & 259.9 & 242.9 & 220.1 & 40.9 & 333.9 & 84.0 \\
    cv2 & 271.0 & 244.9 & 229.3 & 34.9 & 352.7 & 86.7 \\
    cv3 & 268.6 & 251.3 & 237.9 & 33.2 & 327.8 & 83.7 \\
    cv4 & 259.1 & 246.8 & 225.5 & 33.8 & 320.4 & 81.9 \\
    cv5 & 269.7 & 251.0 & 226.4 & 33.1 & 350.0 & 87.9 \\
    Cv std &  5.7 &  3.7 &  6.5 &  3.3 & 14.0 &  2.4 \\
    Cv mean & 265.7 & 247.4 & 227.8 &  35.2 & 337.0 &  84.8 \\
    Test & 268.8 & 252.0 & 227.9 & 34.0 & 344.3 & 83.4 \\
    \midrule
    \textit{Large SimpleBaseline~\cite{DBLP:journals/corr/MartinezHRL17}} & &  & & & & \\
    cv1 & 137.7 & 130.8 & 134.9 & 33.3 & 146.2 & 47.5 \\
    cv2 & 125.5 & 124.9 & 123.6 & 23.1 & 129.1 & 46.0 \\
    cv3 & 126.3 & 124.6 & 125.7 & 19.6 & 128.0 & 44.7 \\
    cv4 & 136.1 & 129.9 & 134.7 & 19.9 & 141.7 & 47.3 \\
    cv5 & 139.0 & 135.7 & 133.4 & 21.4 & 149.8 & 51.2 \\
    Cv std & 6.5 & 4.6 & 5.4 & 5.7 & 9.9 & 2.4 \\
    Cv mean & 132.9 & 129.2 & 130.5 &  23.5 & 139.0 &  47.3 \\
    Test & 131.4 & 131.6 & 120.6 & 19.8 & 148.8 & 44.8 \\
  \midrule
    \textit{CanonPose~\cite{DBLP:journals/corr/abs-2011-14679(canonpose)}} &  & &  &  &  &  \\
    cv1 & 256.7 & 237.1 & 278.9 & 39.1 & 231.4 & 55.1 \\
    cv2 & 255.5 & 244.2 & 284.1 & 35.6 & 215.4 & 56.1 \\
    cv3 & 261.4 & 245.0 & 291.2 & 31.5 & 222.2 & 54.8 \\
    cv4 & 261.3 & 243.4 & 285.5 & 31.6 & 231.7 & 56.8 \\
    cv5 & 270.6 & 250.2 & 292.5 & 35.0 & 246.2 & 61.0 \\
    Cv std &  5.9 &  4.7 &  5.5 &  3.2 & 11.6 &  2.5 \\
    Cv mean & 261.1 & 244.0 & 286.4 &  34.6 & 229.4 &  56.8 \\
    Test & 285.0 & 264.4 & 319.7 & 31.9 & 240.0 & 56.2 \\
    \midrule
    \textit{CanonPose~\cite{DBLP:journals/corr/abs-2011-14679(canonpose)} + 3D sv.} & &  &  &  &  & \\
    cv1 & 163.6 & 155.7 & 160.2 & 33.7 & 173.5 & 49.1 \\
    cv2 & 158.5 & 153.0 & 161.0 & 25.8 & 157.4 & 48.0 \\
    cv3 & 157.9 & 150.0 & 161.5 & 21.8 & 156.5 & 47.3 \\
    cv4 & 157.3 & 154.1 & 155.5 & 22.7 & 162.1 & 49.0 \\
    cv5 & 175.1 & 168.9 & 169.3 & 25.4 & 187.9 & 55.5 \\
    Cv std &  7.5 &  7.3 &  5.0 &  4.7 & 13.3 &  3.3 \\
    Cv mean & 162.5 & 156.3 & 161.5 &  25.9 & 167.5 &  49.8 \\
    Test & 163.6 & 155.9 & 161.3 & 22.2 & 171.4 & 47.4 \\
    \midrule
    \textit{Jointformer~\cite{https://doi.org/10.48550/arxiv.2208.03704(jointformer)} }&  &  &  & &  &  \\
    cv1 & 121.5 & 114.8 & 100.9 & 34.3 & 158.6 & 55.9 \\
    cv2 & 112.5 & 104.9 & 93.4 & 25.2 & 147.5 & 56.6 \\
    cv3 & 110.5 & 101.2 & 94.3 & 20.6 & 141.9 & 56.2 \\
    cv4 & 123.5 & 115.7 & 104.5 & 21.1 & 158.7 & 58.2 \\
    cv5 & 129.4 & 116.0 & 107.9 & 22.5 & 171.6 & 61.1 \\
    Cv std &  7.9 &  7.0 &  6.3 &  5.6 & 11.5 &  2.1 \\
    Cv mean & 119.5 & 110.5 & 100.2 &  24.7 & 155.7 &  57.6 \\
    Test & 109.2 & 103.0 & 82.4 & 19.8 & 155.9 & 53.5 \\
    \bottomrule
    \end{tabular}
    \caption{Results for I2D$\rightarrow$3D task on each 5-fold and test sets. Results are shown for MPJPE metric. All results are pelvis aligned, except $\dagger$ and $\ddagger$ show nose and wrist aligned results for face and hands, respectively. Sv. is supervision. We observe that \textit{CanonPose} fails to generalize to new subject in the test set and performs worse on the test set.}
    \label{tab:baseline_result2}
\end{table}

\begin{table}[tb]
    \scriptsize
    \centering
    \begin{tabular}{ll lllll}
    \toprule
    \multicolumn{1}{c}{method} & \multicolumn{1}{c}{All} &
    \multicolumn{1}{c}{Body} &
    \multicolumn{2}{c}{Face} &
    \multicolumn{2}{c}{Hand} \\
    \multicolumn{3}{c}{} &  & $\dagger$ &   & $\ddagger$ \\
    \midrule
    \textit{SHN~\cite{DBLP:journals/corr/NewellYD16_(stackhourglass)}+SimpleBaseline~\cite{DBLP:journals/corr/MartinezHRL17} }&  & & & &  &\\
    cv1 & 191.0 & 177.9 & 159.8 & 41.4 & 248.7 & 66.1 \\
    cv2 & 159.4 & 151.0 & 135.8 & 30.4 & 202.3 & 62.6 \\
    cv3 & 170.8 & 169.9 & 157.0 & 25.8 & 193.9 & 64.7 \\
    cv4 & 204.8 & 202.3 & 192.9 & 27.7 & 225.5 & 68.0 \\
    cv5 & 204.8 & 192.7 & 173.8 & 30.2 & 261.5 & 71.8 \\
    Cv std & 20.4 & 20.0 & 21.2 &  6.1 & 29.0 &  3.5 \\
    Cv mean & 186.2 & 178.8 & 163.9 &  31.1 & 226.4 &  66.6 \\
    Test & 182.5 & 189.6 & 138.7 & 32.5 & 249.4 & 64.3 \\

    \midrule
        CPN~\cite{DBLP:journals/corr/abs-1711-07319(CPN)}+Jointformer\cite{https://doi.org/10.48550/arxiv.2208.03704(jointformer)} & &  & & & & \\
    cv1 & 100.8 & 101.6 & 75.5 & 29.9 & 141.3 & 53.5 \\
    cv2 & 91.9 & 89.8 & 70.6 & 22.8 & 127.5 & 52.9 \\
    cv3 & 75.7 & 77.5 & 62.5 & 14.9 & 96.0 & 51.0 \\
    cv4 & 78.1 & 82.0 & 58.4 & 16.9 & 107.6 & 52.7 \\
    cv5 & 100.8 & 98.3 & 73.1 & 19.5 & 147.2 & 59.0 \\
    Cv std & 12.1 & 10.3 &  7.3 &  5.9 & 21.8 &  3.0 \\
    Cv mean &  89.5 &  89.8 &  68.0 &  20.8 & 123.9 &  53.8 \\
    Test & 132.6 & 142.8 & 91.9 & 20.7 & 192.7 & 56.9 \\

    \midrule
    
    \textit{Resnet50~\cite{DBLP:journals/corr/HeZRS15(Resnet)}} &  &  & &  & & \\
    cv1 & 123.8 & 117.7 & 97.6 & 34.9 & 169.6 & 58.3 \\
    cv2 & 111.8 & 107.1 & 88.3 & 25.3 & 152.4 & 57.2 \\
    cv3 & 102.5 & 103.8 & 81.9 & 20.0 & 135.0 & 57.8 \\
    cv4 & 113.5 & 114.5 & 89.9 & 21.2 & 151.3 & 58.5 \\
    cv5 & 122.8 & 119.5 & 91.7 & 23.1 & 175.1 & 62.6 \\
    Cv std &  8.8 &  6.8 &  5.78 &  5.9 & 16.0 &  2.1 \\
    Cv mean & 114.9 & 112.5 &  89.9 &  24.9 & 156.7 &  58.9 \\
    Test & 166.7 & 151.6 & 123.6 & 26.3 & 244.9 & 63.1 \\
    \bottomrule
    \end{tabular}
    \caption{Results for RGB$\rightarrow$3D task on each 5-fold and test sets. Results are shown for MPJPE metric. All results are pelvis aligned, except $\dagger$ and $\ddagger$ show nose and wrist aligned results for face and hands, respectively.}
    \label{tab:baseline_result3}
\end{table}

\section{Other issues}
\label{sec:non_technique_issues}
We plan to setup a server for test set evaluation. We will also release the test data after 3-5 years once they are well studied to allow long term use without relying on our evaluation server.

Concerning the license, we only release entirely new labels, which fits the license agreement allowing research output.


\end{document}